\documentclass{theoretics}

\title{Boosting Simple Learners}

\ThCSauthor[princeton]{Noga Alon}{nalon@math.princeton.edu}[0000-0003-1332-4883]
\ThCSauthor[gonenaffil]{Alon Gonen}{alongnn@gmail.com}
\ThCSauthor[hazanaffil]{Elad Hazan}{ehazan@princeton.edu}[0000-0002-1566-3216]
\ThCSauthor[moran]{Shay Moran}{smoran@technion.ac.il}[0000-0002-8662-2737]

\ThCSreceived{Mar 27, 2022}
\ThCSrevised{Jan 2, 2023}
\ThCSaccepted{Mar 20, 2023}
\ThCSpublished{Jun 12, 2023}

\ThCSyear{2023}
\ThCSarticlenum{7}
\ThCSdoi{10.46298/theoretics.23.7}

\ThCSaffil[princeton]{Department of Mathematics, Princeton University.} 
\ThCSaffil[gonenaffil]{OrCam, Israel.}
\ThCSaffil[hazanaffil]{Google AI Princeton and Princeton University.}
\ThCSaffil[moran]{Departments of Mathematics and Computer Science, Technion and Google Research.}

\ThCSthanks{A preliminary version of this work appeared in STOC 2021.

Noga Alon's research is supported in part by NSF grant DMS-1855464 and by a BSF grant 2018267.

Shay Moran is a Robert J.\ Shillman Fellow and supported by ISF grant 1225/20, by BSF grant 2018385, by an Azrieli Faculty Fellowship, by Israel PBC-VATAT, by the Technion Center for Machine Learning and Intelligent Systems (MLIS), and by the the European Union (ERC, GENERALIZATION, 101039692). Views and opinions expressed are however those of the author(s) only and do not necessarily reflect those of the European Union or the European Research Council Executive Agency. Neither the European Union nor the granting authority can be held responsible for them.
}

\usepackage{amsmath} 
\usepackage{framed} 
\usepackage[capitalise,noabbrev,nameinlink]{cleveref}
\ThCSnewtheoita{assumption}  
\ThCSnewtheoita{question}  

\newcommand{\vc}{\mathrm{VC}}
\newcommand{\R}{\mathbb{R}}
\newcommand{\W}{\mathcal{W}}
\newcommand{\X}{\mathcal{X}}
\renewcommand{\H}{\mathcal{H}}
\newcommand{\B}{\mathcal{B}}
\newcommand{\E}{\mathbb{E}}

\newcommand{\A}{\mathcal{A}}
\newcommand{\eps}{\epsilon}
\newcommand{\conv}{\mathsf{CONV}}
\newcommand{\disc}{\mathsf{disc}}
\newcommand{\supp}{\mathsf{supp}}
\newcommand{\sign}{\mathsf{sign}}
\newcommand{\ds}{\mathsf{DS}}
\newcommand{\hs}{\mathsf{HS}}
\newcommand{\w}{\mathsf{w}}

\begin{document}
\maketitle

\begin{abstract}
Boosting is a celebrated machine learning approach which is based on the idea of combining weak and moderately inaccurate hypotheses  
to a strong and accurate one. 
We study boosting under the assumption that the weak hypotheses belong to a class of bounded capacity. This assumption is inspired by the common convention that weak hypotheses are ``rules-of-thumbs'' from an ``easy-to-learn class''. (Schapire and Freund~'12, Shalev-Shwartz and Ben-David '14.) 
Formally, we assume the class of weak hypotheses has a bounded VC dimension. We focus on two main questions:

(i) Oracle Complexity: 
How many weak hypotheses are needed to produce an accurate hypothesis?
We design a novel boosting algorithm and demonstrate that it circumvents a classical lower bound by Freund and Schapire (1995, 2012). 
Whereas the lower bound shows that $\Omega({1}/{\gamma^2})$ weak hypotheses with $\gamma$-margin
are sometimes necessary, our new method requires only $\tilde{O}({1}/{\gamma})$ weak hypothesis,
provided that they belong to a class of bounded VC dimension.
Unlike previous boosting algorithms which aggregate the weak hypotheses by majority votes, the new boosting algorithm uses more complex (``deeper'') aggregation rules. 
We complement this result by showing that complex aggregation rules are in fact necessary to circumvent the aforementioned lower bound.

(ii) Expressivity: Which tasks can be learned by boosting weak hypotheses from a bounded VC class? Can complex concepts that are ``far away'' from the class be learned? 
Towards answering the first question we {introduce combinatorial-geometric parameters which capture expressivity in boosting.} 
As a corollary we provide an affirmative answer to the second question for well-studied classes, including half-spaces and decision stumps.
Along the way, we establish and exploit connections with Discrepancy Theory.

\end{abstract}

\section{Introduction}
Boosting is a fundamental and powerful framework in machine learning which concerns methods 
	for learning complex tasks using combinations of weak learning rules. 
	It offers a convenient reduction approach, whereby in order to learn a given classification task, 
	it suffices to find moderately inaccurate learning rules (called ``{\it weak hypotheses}''), 
	which are then automatically aggregated by the boosting algorithm into an arbitrarily accurate one.
	The weak hypotheses are often thought of as \emph{simple} prediction-rules:

\begin{quote}
``Boosting refers to a general and provably effective method of producing a very accurate prediction rule by combining rough and moderately inaccurate \emph{rules of thumb}.''
\cite[Chapter~1]{Schapire2012}
\end{quote}

\begin{quote}
``\ldots an hypothesis that comes from an \emph{easy-to-learn hypothesis class} and performs just slightly better than a random guess.'' \cite[Chapter~10: Boosting]{Shalev-Shwartz2014}
\end{quote}

In this work we explore how does the simplicity of the weak hypotheses affects the complexity of the overall boosting algorithm:
let $\mathcal{B}$ denote the {\it base-class} which consists of the weak hypotheses used in the boosting procedure.
	For example, $\B$ may consist of all 1-dimensional threshold functions.\footnote{I.e., hypotheses $h:\R\to\{\pm 1\}$ with at most one sign-change.}
	Can one learn arbitrarily complex concepts $c:\R\to\{\pm 1\}$ by aggregating thresholds in a boosting procedure?
	Can one do so by simple aggregation rules such as {\it weighted majority}? 
	How many thresholds must one aggregate to successfully learn a given target concept~$c$?
	How does this number scale with the complexity of $c$?

\paragraph{Target-Class Oriented Boosting (traditional perspective).}
It is instructive to compare the above view of boosting with the traditional perspective. 
	The pioneering manuscripts on this topic (e.g.\ \cite{Kearns88unpublished,Schapire90boosting,Freund90majority}) explored 
	the question of boosting a weak learner in the {\it Probably Approximately Correct} (PAC) setting~\citep{Valiant84PAC}:
	let $\H\subseteq\{\pm 1\}^\X$ be a concept class;
	a {\it $\gamma$-weak learner} for~$\H$ is an algorithm $\W$ which satisfies the following {\it weak learning guarantee}:
	let $c\in \H$ be an arbitrary target concept and let $D$ be an arbitrary target distribution on $\X$.
	(It is important to note that it is assumed here that the target concept $c$ is in $\H$.)
	The input to $\W$ is a confidence parameter $\delta > 0$ and a sample $S$ of $m_0=m_0(\delta)$ examples $(x_i,c(x_i)))$, where the $x_i$'s are drawn independently from $D$.
	The weak learning guarantee asserts that the hypothesis $h=\W(S)$ outputted by $\W$ satisfies
	\[\E_{x\sim D}[h(x)\cdot c(x)] \geq \gamma,\]
	with probability at least $1-\delta$.
	That is, $\W$ is able to provide a non-trivial (but far from desired) approximation to any target-concept $c\in \H$.
	The goal of boosting is to efficiently\footnote{Note that from a {\it sample-complexity} perspective, the task of boosting can be analyzed by basic VC theory:
	by the existence of a weak learner $W$ whose sample complexity is $m_0$, it follows that the VC dimension of $\H$ is $O(m_0(\delta))$ for $\delta=1/2$.
	Then, by the {\it Fundamental Theorem of PAC Learning}, the sample complexity of (strongly) PAC learning $\H$
	is $\tilde O(({d+\log(1/\delta)})/{\eps})$.}
 	convert $\W$ to a strong PAC learner which can approximate $c$ arbitrarily well.
	That is, an algorithm whose input consist of an error and confidence parameters $\eps,\delta>0$
	and a polynomial number of $m(\eps,\delta)$ examples, and whose output is an hypothesis $h'$ such that
	\[\E_{x\sim D}[h'(x)\cdot c(x)] \geq 1-\epsilon,\]	
	with probability at least $1-\delta$.
 	For a text-book introduction see, e.g., \cite[Chapter~2.3.2]{Schapire2012} and \cite[Definition~10.1]{Shalev-Shwartz2014}.

\paragraph{Base-Class Oriented Boosting (this work).}
In this manuscript, we study boosting under the assumption that one first specifies a \emph{fixed} base-class $\mathcal{B}$ 
	of weak hypotheses, and the goal is to aggregate hypotheses from~$\mathcal{B}$
	to learn target-concepts that may be \emph{far-away} from $\mathcal{B}$.
	(Unlike the traditional view of boosting discussed above.)
	In practice, the choice of $\mathcal{B}$ may be done according to prior information on the relevant learning task.
	
Fix a base-class $\mathcal{B}$.
	Which target concepts $c$ can be learned? How ``far-away'' from $\mathcal{B}$ can $c$ be?
	To address this question we revisit the standard weak learning assumption which, in this context, can be rephrased as follows:
	the target concept $c$ satisfies that for every distribution $D$ over $\X$ there exists $h\in\mathcal{B}$ such that
	\[\E_{x\sim D}[h(x)\cdot c(x)] \geq \gamma.\]
	(Notice that the weak learning assumption poses a restriction on the target concept $c$ by requiring it
	to exhibit correlation $\geq \gamma$ with $\mathcal{B}$ with respect to arbitrary distributions.)
	The weak learner $\W$ is given an i.i.d.\ sample of $m_0(\delta)$ random $c$-labelled examples drawn from $D$, 
	and is guaranteed to output an hypothesis $h\in \B$ which satisfies the above with probability at least $1-\delta$.
	In contrast with the traditional ``Target-Class Oriented Boosting'' perspective discussed above, the weak learning algorithm here is
	a \emph{strong} learner for the base-class~$\mathcal{B}$ in the sense that whenever there exists $h\in\mathcal{B}$
	which is $\gamma$-correlated with a target-concept $c$ with respect to a target-distribution~$D$,
	then $\W$ is guaranteed to find such an $h$.
	The weakness of $\mathcal{W}$ is manifested via the simplicity of the hypotheses in $\mathcal{B}$. 

\vspace{2mm}	

This perspective of boosting is common in real-world applications.
For example, the well-studied Viola-Jones object detection framework 
uses simple rectangular-based prediction rules as weak hypotheses for the task of object detection~\citep{Viola01rapidobject}.

\paragraph{Main Questions.}
We are interested in the interplay between the simplicity of the base-class~$\mathcal{B}$  and the expressiveness and efficiency of the boosting algorithm.
    The following aspects will be our main focus:

 \begin{quote}
    \begin{enumerate}
    \item \textbf{Expressiveness:} Given a small edge parameter
      $\gamma>0$, how rich is the class of tasks that can be learned
      by boosting weak hypotheses from~$\mathcal{B}$?  At what
      ``rate'' does this class grow as $\gamma\to 0$?  How about when
      $\B$ is a well-studied class such as {\it Decision stumps} or
      {\it Halfspaces}?
    \item \textbf{Oracle Complexity:} How many times must the boosting
      algorithm apply a weak learner to learn a task which is
      $\gamma$-correlated with $\B$?  Can one improve upon the
      $\tilde O(1/\gamma^2)$ bound which is exhibited by classical
      algorithms such as Adaboost?  Note that each call to the weak
      learner $\mathcal{W}$ amounts to solving an optimization problem
      w.r.t.\ $\B$.  Thus, saving upon this resource can significantly
      improve the overall running time of the algorithm.
    \end{enumerate}
  \end{quote}

The base-class oriented perspective has been considered by previous works such as \cite{Breiman97arcingthe,Friedman00greedyfunction,Mason00boostingalgorithms,Friedman02stochastic,Blanchard03boosting, Lugosi04,Bartlett07consistent,Mukherjee13multiclass}.
These works design specific learning algorithms that are based on aggregating hypotheses from the base-class.
In particular these works remove the weak learner in the sense that the weak hypothesis which is obtained in each round is computed explicitly by optimizing an appropriate function on the data (e.g., maximizing the ``margin'' \citep{Schapire2012} or the ``edge'' \citep{Breiman97arcingthe}). 
In other words, instead of having an oracle access to an arbitrary learner which is only assumed to satisfy the weak learning assumption, these works use carefully tailored way of picking the next weak hypothesis from the base class $B$.
Consequently, the notion of oracle-complexity (which is a central resource in our framework) is irrelevant in these works.
Furthermore, these works focus only on the standard aggregation rule by weighted majority, 
whereas the results in this manuscript exploit the possibility of using more complex rules
and explore their expressiveness.

\paragraph{Outline.}
We begin with presenting the main definitions and results in \Cref{sec:main}: in \Cref{sec:mainoracle} we present a new boosting method whose oracle complexity is only $\tilde{O}({1}/{\gamma})$ weak hypothesis, provided that they belong to a class of bounded VC dimension. We also analyze its generalization performence.
In \Cref{sec:mainexpressivity} we study limits on the expressivity of base-classes; that is, we address the questions which distributions can be learned by boosting an agnostic learner to a given base-class $B$. Towards this end we identify to combinatorial-geometric dimensions called the $\gamma$-VC dimension and $\gamma$-interpolation dimension which provide quantitative bounds on the expressivity.

In \Cref{sec:overview} we overview the main technical ideas used in our proofs, and finally \Cref{sec:oracle} and \Cref{sec:expressivity} contain the proofs:
In \Cref{sec:oracle} we prove the results regarding oracle-complexity, and in \Cref{sec:expressivity} the results regarding expressivity.
Each of \Cref{sec:oracle} and \Cref{sec:expressivity} can be read independently after \Cref{sec:main}
with one exception: the oracle-complexity lower bound in \Cref{sec:oracle}
relies on the theory developed in \Cref{sec:expressivity}.
Finally, \Cref{sec:conc} contains some suggestions for future research.

\section{Main Results}\label{sec:main}

In this section we provide an overview of the main results in this manuscript. 

\paragraph{Weak Learnability.}
Our starting point is a reformulation of the weak learnability assumption in a way which is more suitable to our setting.
	Recall that the $\gamma$-weak learnability assumption asserts that if $c:\X\to \{\pm 1\}$ is the target concept
	then, if the weak learner is given enough $c$-labeled examples drawn from any input distribution over $\X$, 
	it will return an hypothesis which is $\gamma$-correlated with $c$. 
	Since here it is assumed that the weak learner is a strong learner for the base-class $\B$,
	one can rephrase the weak learnability assumption only in terms of~$\B$ using the following notion\footnote{In fact, $\gamma$-realizability corresponds to the {\it empirical weak learning assumption} by \cite[Chapter~2.3.2]{Schapire2012}. The latter is a weakening of the standard weak PAC learning assumption which suffices to guarantee generalization.}:
\begin{definition}[$\gamma$-realizable samples/distributions] \label{def:gammaRealizable}
Let $\B\subseteq\{\pm 1\}^\X$ be the base-class, let $\gamma\in(0,1)$. 
    A sample~$S=((x_1,y_1),\ldots,(x_m,y_m))$ is \emph{$\gamma$-realizable} with respect to $\mathcal{B}$
    if for any probability distribution $Q$ over $S$ there exists $b\in\mathcal{B}$ such that 
\[
 \mathsf{corr}_Q(b) := \E_{(x,y)\sim Q}[b(x)\cdot y] \ge \gamma.
\]
We say that a distribution $D$ over $\mathcal{X}\times\{\pm 1\}$ is {\it $\gamma$-realizable} if any i.i.d.\ sample drawn from~$D$ is $\gamma$-realizable.\footnote{We note that one can relax the definition of $\gamma$-realizable distribution by requiring that a random sample from it is $\gamma$-realizable w.h.p.\ (rather than w.p.\ $1$). Consequently, the results in this paper which use this definition also hold w.h.p. However, for the sake of exposition we work with the above definition.}
\end{definition}
Thus, the $\gamma$-weak learnability assumption boils down to assuming that the target distribution is $\gamma$-realizable.

Note that for $\gamma=1$ the notion of $\gamma$-realizability specializes to the classical notion of realizability (i.e., consistency with the class). 
	Also note that as $\gamma\to 0$, the set of $\gamma$-realizable samples becomes larger.

\paragraph{Quantifying Simplicity.}
Inspired by the common intuition that weak hypotheses are ``rules-of-thumb'' \citep{Schapire2012} that belong to an ``{easy-to-learn} hypothesis class''~\citep{Shalev-Shwartz2014},
	we  make the following assumption:
 
\begin{assumption}[Simplicity of Weak Hypotheses]
	Let $\B\subseteq \{\pm 1\}^\X$ denote the base-class which contains the weak hypotheses
	provided by the weak learner. Then, $\B$ is a VC class; that is, $\vc(\B)=O(1)$.
\end{assumption}

\subsection{Oracle Complexity (Section~\ref{sec:oracle})}
\label{sec:mainoracle}

\subsubsection{Upper Bound (Section~\ref{sec:oracleub})}
Can the assumption that $\B$ is a VC class be utilized
	to improve upon existing boosting algorithms?
	We provide an affirmative answer by using it to circumvent a classical lower bound on the oracle-complexity of boosting. 
	Recall that the oracle-complexity refers to the number of times the boosting algorithm calls the weak learner during the execution.
	As discussed earlier, it is an important computational resource and it controls a cardinal part of the running time of classical boosting algorithms such as Adaboost.

\paragraph{A Lower Bound by \cite{Freund90majority} and  \cite[Chapter~13.2.2]{Schapire2012}.}
	Freund and Schapire showed that for any fixed edge parameter $\gamma$,  every boosting procedure must invoke the weak learner
	at least $\Omega(1/\gamma^2)$ times in the worst-case.
	That is, for every boosting algorithm $\A$ and every $\gamma>0$ 
	there exists a $\gamma$-weak learner $\W=\W(\A,\gamma)$ and a target distribution such that
	 $\A$ must invoke $\W$ at least $\Omega(1/\gamma^2)$ times in order to obtain a constant population loss, say $\leq1/10$
	 \cite[Chapter~13.2.2]{Schapire2012}.

However, the ``bad'' weak learner $\W$ is constructed using a probabilistic argument;
	in particular the VC dimension of the corresponding base-class of weak hypotheses is $\omega(1)$.
	Thus, this result leaves open the possibility of achieving an $o(1/\gamma^2)$ oracle-complexity,
	under the assumption that the base-class $\B$ is a VC class.

 We demonstrate a boosting procedure called {\it Graph Separation Boosting} (\Cref{alg:sepBoost}) which,
	under the assumption that $\B$ is a VC class, invokes the weak learner only $\tilde O(\frac{\log(1/\epsilon)}{\gamma})$ times and achieves generalization error~$\leq\eps$.    
	We stress that \Cref{alg:sepBoost} is oblivious to the advantage parameter $\gamma$ and to the class $\B$.	
	(I.e., it does not not ``know'' $\B$ nor $\gamma$.)
	The assumption that $\B$ is a VC class is only used in the analysis.

It will be convenient in this part to weaken the weak learnability assumption as follows:
	for any $\gamma$-realizable distribution $D$, if $\W$  is fed with a sample $S'\sim D^{m_0}$
	then $\E_{S'\sim D^{m_0}}\bigl[\mathsf{corr}_D\bigl(\W(S')\bigr)\bigr]\geq \gamma/2$.
	That is, we only require that \emph{expected} correlation of the output hypothesis is at least $\gamma/2$
	(rather than with high probability).

\begin{algorithm}
\SetKwComment{Comment}{}{}
\SetKwInput{Parameters}{Parameters} 
\Parameters{a base-class $\B$, a weak learner $\W$ with sample complexity $m_0$, an advantage parameter $\gamma>0$}

\SetKwInput{WL}{Weak Learnability} 
\WL{for every distribution $D$ which is $\gamma$-realizable by $\B$: $\E_{S'\sim D^{m_0}}\bigl[\mathsf{corr}_D\bigl(\W(S')\bigr)\bigr]\geq \gamma/2$}

\KwIn{a sample $S=((x_{1},y_1),\ldots,(x_{m},y_m))$ which is $\gamma$-realizable by $\B$, and a black-box oracle access to the weak learner $\W$.}
Define an undirected graph $G=(V,E)$ where $V=[m]$ and $\{i,j\}\in E\Leftrightarrow y_{i}\neq y_{j}$.\;
Set $t\leftarrow0$\;
\While{$E\neq\emptyset$}{ 
$t:=t+1$\; 
Define distribution $P_{t}$ on $S:~ P_{t}(x_i,y_i)\propto deg_G(i)$.\;
\tcp*[r]{($\deg_G(\cdot )$ is the degree in the graph $G$.)}
Draw a sample $S_t \sim P_{t}^{m_0}$\;
Set $b_{t}\leftarrow\mathcal{A}(S_t)$\;
Remove from $E$ every edge $\{i,j\}$ such that $b_{t}(x_{i})\neq b_{t}(x_{j})$\;
}
Set $T\leftarrow t$\;
Compute an aggregation rule $f:\{\pm 1\}^T\to\{\pm 1\}$  such that the aggregated hypothesis $f(b_{1},\ldots b_{T})$ is consistent with $S$.\label{line13}\;
\tcp*[r]{($f$ exists by \Cref{lem:sep}.)}
Output $\hat{h}=f(b_{1},\ldots,b_{T})$.
\caption{Graph Separation Boosting}
\label{alg:sepBoost}
\end{algorithm}

The main idea guiding the algorithm is quite simple. 
	We wish to collect as fast as possible a set of weak hypotheses $b_{1},\ldots,b_{T}\in\mathcal{B}$
	that can be aggregated into a {\it consistent hypothesis}.
	That is, a hypothesis $h\in\{\pm 1\}^{X}$ of the form 
	\[h=f(b_1,\ldots, b_T),\]
	for some aggregation rule $f:\{\pm 1\}^T \to \{\pm 1\}$ such that $h(x_{i})=y_{i}$ for all examples $(x_i,y_i)$ in the input sample $S$.
	An elementary argument shows that such an $h$ exists if and only if for every pair of examples $(x_i,y_i),(x_j,y_j)\in S$ of opposite labels (i.e., $y_i\neq y_j$)
	there is a weak hypothesis that separates them. That is,
	\[(\forall y_i\neq y_j) (\exists b_k) : b_k(x_i)\neq b_k(x_j).\]
	The algorithm thus proceeds by greedily reweighing the examples in $S$
	in way which maximizes the number of separated pairs.

The following theorem shows that the (expected) number of calls to the weak learner until all pairs are separated is some $T=O(\log(\lvert S\rvert)/\gamma)$.
	The theorem is stated in terms of the number of rounds, 
	but as the weak learner is called one time per round, 
	the number of rounds is equal to the oracle-complexity.
\begin{theorem} [Oracle Complexity Upper Bound]  \label{thm:mainOracleComp}
Let $S$ be an input sample of size $m$ which is $\gamma$-realizable with respect to $\B$, 
and let $T$ denote the number of rounds \Cref{alg:sepBoost} performs when applied on~$S$. 
Then, for every~$t\in\mathbb{N}$ 
\[\Pr[T\geq t] \leq \exp\bigl(2\log m -t \gamma/2\bigr).\]
In particular, this implies that $\E[T]=O(\log (m)/\gamma)$.
\end{theorem}

\paragraph{Generalization Bounds (\Cref{sec:gen}).}
An important subtlety in \Cref{alg:sepBoost} is that it does not specify how to find the aggregation rule $f$  in Line~\ref{line13}.
	In this sense, \Cref{alg:sepBoost} is in fact a meta-algorithm.

	It is possible that for different classes $\B$ one can implement Line~\ref{line13} 
 	in different ways which depend on the structure of $\B$ and yields favorable rules $f$.\footnote{For example, when $\B$ is the class of one dimensional thresholds, see  \Cref{sec:oracleub}.} 
	In practice, one might also consider applying heuristics to find~$f$:
	e.g., consider the $T=O(\log m/\gamma)$ dimensional representation $x_i\mapsto (b_1(x_i),\ldots,b_T(x_i))$ which is implied by the weak hypotheses, 
	and train a neural network to find an interpolating rule $f$.\footnote{Observe in this context that the common weighted-majority-vote aggregation rule can be viewed  as a single neuron with a threshold activation function.} 
	(Recall that such an $f$ is guaranteed to exist, since $b_1,\ldots,b_T$ separate all opposite-labelled pairs.)

To accommodate the flexibility in computing the aggregation rule in Line~\ref{line13}, 
	we provide a generalization bound which  {\it adapts to complexity of the aggregation rule}.
	That is, a bound which yields better generalization guarantees for simpler rules.
	Formally, we follow the notation in \cite[Chapter~4.2.2]{Schapire2012} and assume that for every sequence of weak hypotheses $b_1\ldots b_T\in\B$
	there is an {\it aggregation class} 
	\[\H = \H(b_1,\ldots, b_T)\subseteq \Bigl\{ f(b_1\ldots b_T) : f:\{\pm 1\}^T\to \{\pm 1\}\Bigr\},\]
	such that the output hypothesis of \Cref{alg:sepBoost} is a member of $\H$. 
	For example, for classical boosting algorithms such as Adaboost, $\H$ is the class of all weighted majorities
	$\{\sign(\sum_i w_i\cdot b_i) : w_i\in\R\}$, and the particular weighted majority in $\H$ which is outputted depends
	on the input sample $S$.

\begin{theorem}[Aggregation-Dependent Bounds]\label{prop:genalg}
\label{thm:mainGeneralizationAnyBoost}
Assume that the input sample $S$ to \Cref{alg:sepBoost} is drawn from a distribution $D$ which is $\gamma$-realizable with respect to $\B$.
Let $b_1\ldots b_T$ denote the hypotheses outputted by $\W$ during the execution of $\Cref{alg:sepBoost}$ on $S$,
and let $\H=\H(b_1\ldots b_T)$ denote the aggregation class.
Then, the following occurs with probability at least $1-\delta$:
\begin{enumerate}
\item{\bf Oracle Complexity}: the number of times the weak learner is called satisfies
\[T=O\Bigl(\frac{\log m + \log(1/\delta)}{\gamma}\Bigr).\]
\item {\bf Sample Complexity}: the hypothesis $h\in\H$ outputted by \Cref{alg:sepBoost} satisfies $\mathsf{corr}_D(h) \geq 1-\eps$, where
\[
\eps = O \left (\frac{\bigl(T\cdot m_0 + \vc(\H)\bigr)\log m+\log(1/\delta)  }{m} \right) = \tilde O\Bigl(\frac{m_0 }{\gamma\cdot m} + \frac{\vc(\H)}{m}\Bigr),
\]
where $m_0$ is the sample complexity of the weak learner $\W$.
\end{enumerate}
\end{theorem}

\Cref{prop:genalg} demonstrates an upper bound on both the oracle and sample complexities of \Cref{alg:sepBoost}.
The sample complexity upper bound is algorithm-dependent in the sense that it depends on $\vc(\H)$
the VC dimension of $\H=\H(b_1\ldots b_T)$ -- the class of possible aggregations outputted by the algorithm.
In particular $\vc(\H)$ depends on the base-class $\B$ and on the implementation of Line \ref{line13} in \Cref{alg:sepBoost}.
Notice that the class $\H(b_1\ldots b_T)$ is \emph{data-dependent}: it is a function of the input sample of the algorithm. 
Thus, the generalization bound above does not follow from standard VC generalization bounds that apply for 
fixed (and data-independent) classes.
The way we control this data dependency is via the notion of \emph{hybrid sample compression schemes} \citep{Schapire2012};
recall that in standard sample compression schemes, the output hypothesis is a function of a (small) subset of the training examples.
Hybrid sample compression schemes are an extension of sample compression schemes 
in which the output hypothesis is instead selected from a class of hypotheses $\H$,
where the class (rather than the hypothesis itself) is a function of a (small) subset of the data.
See Section~\ref{sec:gen} for more details.

\medskip

How large can $\vc(\H)$ be for a given class of simple aggregation rules?
	The following combinatorial proposition addresses this question quantitatively.
	Here, it is assumed the aggregation rule used by \Cref{alg:sepBoost} belong to a fixed class $G$ of ``$\{\pm 1\}^T\to\{\pm 1\}$'' functions.
	For example, $G$ may consist of all weighted majority votes $g(x_1,\ldots,x_T) = \sign(\sum w_i\cdot x_i)$, for $w_i\in \R$,
	or of all networks with of some prespecified topology and activation functions, etcetera.

\begin{proposition}[VC dimension of Aggregation]\label{prop:vcagr}
Let $\B \subseteq \{\pm 1\}^\X$ be a base-class and let $G$ denote a class of ``$\{\pm 1\}^T\to\{\pm 1\}$'' functions (``aggregation-rules''). Then,
\[\vc\Bigl(\Bigl\{g(b_1,\ldots,b_T) \vert b_i\in \B, g\in G\Bigr\}\Bigr) \leq  c_T\cdot(T\cdot \vc(\B) + \vc(G)),\]
where $c_T = O(\log T)$. 
Moreover, even if $G$ contains all ``$\{\pm 1\}^T\to \{\pm 1\}$'' functions,
	then the following bound holds for every fixed $b_1,b_2,\ldots, b_T\in \B$
\[
\vc\Bigl(\Bigl\{g(b_1,\ldots,b_T) \vert  g:\{\pm 1\}^T\to\{\pm 1\}\Bigr\}\Bigr)
\leq \binom{T}{\leq d^*} 
\leq (eT/d^*)^{d^*}\,,
\]
where $d^*$ is the dual VC dimension of $\B$.
\end{proposition}
So, for example if $G$ consists of all possible majority votes then $\vc(G)\leq T+1$ (because $G$ is a subclass of $T$-dimensional halfspaces),
and $\vc(\H(b_1\ldots b_T))= O(\vc(\B)\cdot T\log T) = \tilde O({\vc(\B)}/{\gamma})$.

\Cref{prop:vcagr} generalizes a result by \cite{Blumer89VC} who considered the case when $G=\{g\}$ consists of a single function.
(See also~\cite{Eisentat07kfold,Csikos19tight}).
In \Cref{sec:oracle} we state and prove \Cref{prop:vccomp} which gives an even more general bound which allows the $b_i$'s to belong to different classes $\B_i$'s.

Note that even if  \Cref{alg:sepBoost} uses arbitrary aggregation rules, 
	\Cref{prop:vcagr} still provides a bound of $\vc(\H(b_1\ldots b_T))\leq(eT/d^*)^{d^*}$, where $d^*$ is the dual VC dimension of $\B$.
	In particular, since $\B$ has VC dimension $d=O(1)$ then also its dual VC dimension satisfies $d^*=O(1)$ and we get a polynomial bound on the complexity of \Cref{alg:sepBoost}:\footnote{In more detail $d^*\leq 2^{d+1}-1$, and for many well-studied classes (such as halfspaces) the VC dimension and its dual are polynomially related~\citep{Assouad83densite}. }
\begin{corollary} \label{thm:mainGeneralization}
Let $\B$ be the base-class, let $d^*$ denote its dual VC dimension, and assume oracle access to a $\gamma$-learner for $\B$
with sample complexity $m_0$.
Assume the input sample~$S$ to \Cref{alg:sepBoost} consists of $m$ examples drawn independently from a $\gamma$-realizable distribution.
Then with probability $1-\delta$ the following holds:
\begin{enumerate}
\item{\bf Oracle Complexity}: the number of times the weak learner is called is $T=O(\frac{\log m + \log(1/\delta)}{\gamma})$.
\item {\bf Sample Complexity}: The hypothesis $h\in\H$ outputted by \Cref{alg:sepBoost} satisfies $\mathsf{corr}_D(h) \geq 1-\eps$, where
\[
\eps = O \left (\frac{\bigl(T\cdot m_0 + T^{d^*}\bigr)\log m+\log(1/\delta)  }{m} \right) = \tilde O\Bigl(\frac{m_0 }{\gamma\cdot m} + \frac{1}{\gamma^{d^*}\cdot m}\Bigr).
\]

\end{enumerate}
\end{corollary}
	This shows that indeed the impossibility result by \cite{Schapire2012} is circumvented when $\B$ is a VC class:
	indeed, in this case the sample size $m$ is bounded by a polynomial function of $1/\epsilon,1/\delta$.
	Note however that obtained generalization bound is quite pessimistic (exponential in $d^*$) and 
	thus, we consider this polynomial bound interesting only from a purely theoretical perspective:
	it serves as a proof of concept that improved guarantees are provably possible when the base-class $\B$ is simple.
	We stress again that for specific classes $\mathcal{B}$ one can come up with explicit and simple aggregation rules 
	and hence obtain better generalization bounds via \Cref{prop:genalg}. 
	We refer the reader to  \Cref{sec:oracle} for a more detailed discussion and the proofs.

\subsubsection{Oracle Complexity Lower Bound (Section~\ref{sec:oraclelb})}
Given that virtually all known boosting algorithms use majority-votes to aggregate the weak hypotheses,
	it is natural to ask whether the $O(1/\gamma)$ oracle-complexity upper bound 
	can be attained if one restricts to aggregation by such rules. 
	We prove an impossibility result, which shows that a nearly quadratic lower bound holds 
	when $\B$ is the class of halfspaces in $\R^d$.

\begin{theorem}[Oracle Complexity Lower Bound]\label{thm:lowerOracleHalfspaces}
Let $\gamma>0$ be the edge parameter, and let $\B=\hs_d$ be the class of $d$-dimensional halfspaces.
	Let $\A$ be a boosting algorithm which uses a (possibly weighted) majority vote as an aggregation rule.
	That is, the output hypothesis of $\A$ is of the form
	\[h(x)=\sign\bigl( w_1\cdot b_1(x) + \ldots + w_T\cdot b_T(x) \bigr),\]
	where $b_1\ldots b_T$ are the weak hypotheses returned by the weak learner, and $w_1,\ldots w_T\in\mathbb{R}$.
	Then, for every weak learner $\W$ which outputs weak hypotheses from $\hs_d$ there exists a distribution $D$
	which is $\gamma$-realizable by $\hs_d$ such that if $\A$ is given sample access to $D$ and oracle access to $\W$, 
	then it must call $\W$ at least
	\[T=\tilde\Omega_d\Bigl(\frac{1}{\gamma^{2-\frac{2}{d+1}}}\Bigr)\]
	times in order to output an hypothesis $h$ such that with probability at least $1-\delta=3/4$ it satisfies $\mathsf{corr}_D(h)\geq 1-\eps=3/4$.
	The $\tilde \Omega_d$ above conceals multiplicative factors which depend on $d$ and logarithmic factors
	which depend on $1/\gamma$.
\end{theorem}

Our proof of \Cref{thm:lowerOracleHalfspaces} is based on a counting argument which applies more generally;
	it can be used to provide similar lower bounds as long as the family of allowed aggregation rules is sufficiently restricted
	(e.g., aggregation rules that can be represented by a bounded circuit of majority-votes, etc).

\subsection{Expressivity (Section~\ref{sec:expressivity})}\label{sec:mainexpressivity}
We next turn to study the expressivity of VC classes as base-classes in the context of boosting.
That is, given a class $\B$, what can be learned using oracle access to a learning algorithm~$\W$ for~$\B$?

It will be convenient to assume that $\B\subseteq\{\pm 1\}^\X$ is {\it symmetric}: 
	\[(\forall b\in\{\pm 1\}^\X): b\in \B \iff -b\in \B.\]
	This assumption does not compromise generality because a learning algorithm for $\B$
	can be converted to a learning algorithm for $\{\pm b : b\in \B\}$ with a similar sample complexity.
	So, if $\B$ is not symmetric, we can replace it by $\{\pm b : b\in \B\}$.

Our starting point is the following proposition, which asserts that under a mild condition,
	any base-class $\B$ can be used via boosting to learn arbitrarily complex tasks as $\gamma\to 0$.

\begin{proposition}[A Condition for Universality]\label{prop:sufficientRank}
The following statements are equivalent for a symmetric class~$\mathcal{B}$:
\begin{enumerate}
    \item For every $c:\X\to\{\pm 1\}$ and every sample $S$ labelled by $c$, there is $\gamma>0$
    such that~$S$ is $\gamma$-realizable by $\mathcal{B}$.
    \item For every $\{x_1,\ldots,x_n\} \subseteq \mathcal{X}$, the linear-span of $\{(b(x_1),\ldots,b(x_n))\in\mathbb{R}^n:~b\in \mathcal{B}\}$ is $n$-dimensional.
\end{enumerate}
\end{proposition}
Item 1 implies that in the limit as $\gamma\to 0$, \emph{any} sample $S$
	can be interpolated by aggregating weak hypotheses from $\B$ in a boosting procedure.
	Indeed, it asserts that any such sample satisfies the weak learning assumption for some $\gamma>0$
	and therefore given oracle access to a sufficiently accurate learning algorithm for $\B$, any boosting algorithm
	will successfully interpolate $S$.

Observe that every class $\mathcal{B}$ that contains singletons or one-dimensional thresholds satisfies Item 2
	and hence also Item 1. Thus, virtually all standard hypothesis classes that are considered in the literature satisfy it.	

It is worth mentioning here that an ``infinite'' version of \Cref{prop:sufficientRank} has been established for some \emph{specific} boosting algorithms.
	Namely, these algorithms have been shown to be  {\it universally consistent} in the sense that 
	their excess risk w.r.t.\ the {\it Bayes optimal} classifier tends to zero in the limit, as the number of examples tends to infinity.
	See e.g.~\cite{Breiman00,Mannor00,Mannor02,Buhlmann03,Jiang04,Lugosi04,Zhang04,Bartlett07consistent}.

\subsubsection{Measuring Expressivity of Base-Classes}

\Cref{prop:sufficientRank} implies that, from a qualitative perspective, any reasonable class
	can be boosted to approximate arbitrarily complex concepts, provided that $\gamma$ is sufficiently small.
	From a realistic perspective, it is natural to ask how small should $\gamma$ be in order to ensure a satisfactory level of expressivity. 

\begin{question}
 Given a fixed small $\gamma>0$, what are the tasks that can be learned by boosting 
	a $\gamma$-learner for $\B$? At which rate does this class of tasks grow as $\gamma\to 0$?
\end{question}

To address this question we propose two combinatorial parameters called the $\gamma$-VC dimension
    and the $\gamma$-interpolation dimension which quantify the size/richness of the family of tasks that can be learned by aggregating hypotheses from $\B$.

\begin{definition}[$\gamma$-interpolation]
Let $\mathcal{B}$ be a class and $\gamma\in[0,1]$ be an edge parameter. 
We say that a set $\{x_{1},\ldots,x_{d}\}\subseteq\mathcal{X}$ is $\gamma$-interpolated by $\mathcal{B}$
if for any $c:\mathcal{X}\rightarrow\{\pm 1\}$, the sample $S=((x_{1},c(x_{1})),\ldots,(x_{d},c(x_{d}))$ 
is $\gamma$-realizable with respect to $\mathcal{B}$. 
\end{definition}

Intuitively, when picking a base-class $\mathcal{B}$,
    one should minimize the VC dimension
    (because then the weak-learning task is easier, and hence each call to the weak learner is less expensive),
    while maximizing the family of $\gamma$-interpolated sets (because then the overall boosting algorithm can learn more complex tasks).
    This gives rise to the following definition, which has been introduced by~\citet*{ChenControl21}.

\begin{definition}[$\gamma$-interpolation dimension]\label{def:gammaid}
Let $\mathcal{B}$ be a class and $\gamma\in[0,1]$ be an edge parameter. 
The $\gamma$-interpolation dimension of $\mathcal{B}$, denoted $\mathrm{ID}_{\gamma}(\mathcal{B})$, 
is the maximal integer $d\geq 0$ for which every subset of~$\mathcal{X}$ of size $d$ is $\gamma$-interpolated.
If $\mathcal{B}$ $\gamma$-interpolates every finite subset of~$\mathcal{X}$ then its $\gamma$-interpolation dimension is defined to be $\infty$.
\end{definition}

We note that this definition might be too restrictive in natural scenarios where it is impossible to $\gamma$-interpolate certain small degenerate sets. 
For example, consider a learning task where $\mathcal{X}=\mathbb{R}^d$ and $\mathcal{B}$ is some geometrically defined class. In such cases, it might be more natural to quantify only over $\gamma$-interpolated sets that are in general position.
Indeed, our results below regarding the expressiveness of half-spaces and decision-stumps
are based on such relevant assumptions.

The following definition extends the classical VC dimension:
\begin{definition}[$\gamma$-VC dimension]\label{def:gammavc}
Let $\mathcal{B}$ be a class and $\gamma\in[0,1]$ be an edge parameter. 
The $\gamma$-VC dimension of $\mathcal{B}$, denoted $\mathrm{VC}_{\gamma}(\mathcal{B})$, 
is the maximal size of a set which is $\gamma$-interpolated by $\mathcal{B}$.
If $\mathcal{B}$ $\gamma$-interpolates sets of arbitrarily large size then its $\gamma$-VC dimension
is defined to be $\infty$.
\end{definition}
Note that for $\gamma=1$, the $\gamma$-VC dimension specializes to the VC dimension,
    which is a standard parameter for measuring the complexity of learning a target concept $c\in\mathcal{B}$.
    Thus, the $\gamma$-VC dimension can be thought of as an extension of the VC dimension to the $\gamma$-realizable 
    setting, where the target concept $c$ is not in $\B$ and it is only $\gamma$-correlated with $\mathcal{B}$.

\begin{observation}
For every class $\mathcal{B}$ and for every $\gamma\in (0,1)$:
\[\mathrm{ID}_{\gamma}(\mathcal{B})\leq \mathrm{VC}_{\gamma}(\mathcal{B}).\]
\end{observation}

\paragraph{General Bounds on the $\gamma$-VC Dimension.}

It is natural to ask how large can the $\gamma$-VC dimension as a function of the VC dimension and $\gamma$.

\begin{theorem}
\label{thm:generalGammaVC}
Let $\mathcal{B}$ be a class with VC dimension $d$.
    Then, for every~$0<\gamma\leq 1$:
    \[\mathrm{VC}_{\gamma}(\mathcal{\mathcal{B}})=O\left(\frac{d}{\gamma^{2}}\log(d/\text{\ensuremath{\gamma}})\right) = \tilde O\Bigl(\frac{d}{\gamma^2}\Bigr).\]
    Moreover, this bound is nearly tight as long as $d$ is not very small compared to $\log(1/\gamma)$: 
    for every $\gamma>0$ and $s\in\mathbb{N}$ there is a class $\mathcal{B}$ of VC dimension $d = O(s\log(1/\gamma))$
    and 
    \[\mathrm{VC}_{\gamma}(\mathcal{B})=\Omega\left(\frac{s}{\gamma^{2}}\right) = \tilde \Omega\Bigl(\frac{d}{\gamma^2}\Bigr).\]
\end{theorem}
Thus, the fastest possible growth of the $\gamma$-VC dimension is asymptotically $\approx d/\gamma^2$.
	We stress that the upper bound here implies an impossibility result; it poses a restriction on the class of tasks that can be approximated
	by boosting a~$\gamma$-learner for $\B$. 

Note that the above lower bound is realized by a class $\B$ whose VC dimension is at least $\Omega(\log(1/\gamma))$,
	which deviates from our focus on the setting where the VC dimension is a constant and $\gamma\to 0$.
	Thus, we prove the next theorem which provides a sharp, subquadratic, dependence on $\gamma$ (but a looser dependence on $d$). 
\begin{theorem}[$\gamma$-VC dimension: improved bound for small $\gamma$]\label{thm:ubdisc}
Let $\mathcal{B}$ be a class with VC dimension $d\ge 1$. Then, for every~$0<\gamma\leq 1$:
\[
\mathrm{VC}_{\gamma}(\mathcal{B}) \leq O_d\left( \left(\frac{1}{\gamma}\right)^{\frac{2d}{d+1}} \right),
\]
where $O_d(\cdot)$ conceals a multiplicative constant that depends only on $d$.
Moreover, the above inequality applies for any class $\B$ whose primal shatter function\footnote{The {\it primal shatter function} of a class $\B\subseteq\{\pm 1\}^\X$
is the minimum $k$ for which there exists a constant $C$ such that for every finite $A\subseteq \X$, the size of $\B|_A = \{b|_A : b\in \B\}$ is at most $C\cdot \lvert A\rvert^k$.
Note that by the Sauer--Shelah--Perles Lemma,  the primal shatter function is at most the VC dimension.} is at most $d$.
\end{theorem}
As we will prove in \Cref{thm:gammaVCHS}, the dependence on $\gamma$ in the above bound is tight. 
	It will be interesting to determine tighter bounds in terms of $d$.

\paragraph{Bounds for Popular Base-Classes.}
We next turn to explore the $\gamma$-VC and $\gamma$-ID dimensions of two well studied geometric classes: halfspaces and decision stumps.

Let $\mathsf{HS}_d$ denote the class of halfspaces (also known as linear classifiers) in $\mathbb{R}^d$.
	That is $\mathsf{HS}_d$ contains all concepts of the form ``$x\mapsto \sign(w\cdot x + b)$'', 
	where $w\in\R^d$, $b\in\R$, and $w\cdot x$ denotes the standard inner product between $w$ and $x$.
	This class is arguably the most well studied class in machine learning theory, and it provides the building blocks
	underlying modern algorithms such as { Neural Networks} and {Kernel Machines}.
	For $\mathsf{HS}_d$ we give a tight bound on its $\gamma$-VC dimesion (in terms of $\gamma$) 
    of $\Theta_d\left(\frac{1}{\gamma}\right)^{\frac{2d}{d+1}}$.
    The upper bound follows from \Cref{thm:ubdisc} and the lower bound is established in the next theorem:
\begin{theorem}[Halfspaces]\label{thm:gammaVCHS}
Let $\mathsf{HS}_d$ denote the class of halfspaces in $\mathbb{R}^d$ and $\gamma\in(0,1]$. 
Then, 
\[\mathrm{VC}_{\gamma}(\mathsf{HS}_d)=\Theta_d\left(\left (\frac{1}{\gamma}\right)^{\frac{2d}{d+1}} \right).\]
Further, every set $Y\subseteq \R^d$ of size $\Theta_d((\frac{1}{\gamma})^{\frac{2d}{d+1}} )$ is $\gamma$-interpolated by $\mathsf{HS}_d$, provided that $Y$ is \emph{dense} in the following sense:
the ratio between the maximal and minimal distances among all distinct pairs of points in $Y$ is bounded by some
$O_d(\lvert Y\rvert^{\frac{1}{d}})$.
\end{theorem}
Thus the class of halfspaces is rather expressive as a base-class; note that natural point sets such as grids
are dense and hence meet the condition for being $\gamma$-interpolated by halfspaces.

We next study the $\gamma$-VC and $\gamma$ ID dimensions of the class of {\it Decision Stumps}.
A $d$-dimensional decision stump is a concept of the form $\sign(s(x_{j}-t))$, where $j\leq d$, $s\in\{\pm 1\}$ and $t\in\R$.
In other words, a decision stump is a halfspace which is aligned with one of the principal axes.
This class is popular in the context of boosting, partially because it is easy to learn it, even in the agnostic setting.
Also note that the Viola-Jones framework hinges on a variant of decision stumps~\citep{Viola01rapidobject}.

\begin{theorem}[Decision Stumps]\label{thm:gammaVCDS}
Let $\mathsf{DS}_d$ denote the class of decision stumps in $\mathbb{R}^d$ and $\gamma\in(0,1]$. 
Then, 
\[\mathrm{VC}_{\gamma}(\mathsf{DS}_d)=O\left( \frac{d}{\gamma}\right).\]
Moreover, the dependence on $\gamma$ is tight, already in the 1-dimensional case.
In fact, for every~$\gamma$ such that $1/\gamma\in \mathbb{N}$
\[\mathsf{ID}_\gamma(\ds_1)\geq  1/\gamma.\] 
For $d>1$, the class of $d$-dimensional decision-stumps $\gamma$-interpolates every set $Y\subseteq \mathbb{R}^d$ of size $1/\gamma$, 
provided that there exists $i\leq d$ so that every pair of distinct points $x,y\in Y$ satisfy $x_i\neq y_i$.
\end{theorem}

Thus, the class of halfspaces exhibits a near quadratic dependence in $1/\gamma$ (which, by \Cref{thm:ubdisc}, is the best possible),
	and the class of decision stumps exhibits a linear dependence in $1/\gamma$.
	In this sense, the class of halfspaces is considerably more expressive.
	On the other hand the class of decision stumps can be learned more efficiently in the agnostic setting,
	and hence the weak learning task is easier with decision stumps.
	
Along the way of deriving the above bounds, 
	we analyze the $\gamma$-VC dimension of one-dimensional classes
	and of unions of one-dimensional classes.
	From a technical perspective,
	we exploit some fundamental results in discrepancy theory.

\section{Technical Overview}\label{sec:overview}

In this section we overview the main ideas which are used in the proofs.
	We also try to guide the reader on which of our proofs reduce to known arguments and which require new ideas.

\subsection{Oracle Complexity}

\subsubsection{Lower Bound}
We begin with overviewing the proof of \Cref{thm:lowerOracleHalfspaces}, which asserts that any
boosting algorithm which uses a (possibly weighted) majority vote
as an aggregation rule is bound to call the weak learner at least nearly~$\Omega(\frac{1}{\gamma^2})$ times,
even if the base-class has a constant VC dimension. 

It may be interesting to note that from a technical perspective, this proof bridges the two parts of the paper.
In particular, it relies heavily on \Cref{thm:gammaVCHS} which bounds the $\gamma$-VC dimension of halfspaces.

The idea is as follows: let $T=T(\gamma)$ denote the minimum number of times 
a boosting algorithm calls a $\gamma$-learner for halfspaces in order to achieve a constant population loss, say $\eps=1/4$.
We show that unless~$T$ is sufficiently large (nearly quadratic in $\frac{1}{\gamma}$), 
then there must exists a $\gamma$-realizable learning task (i.e., which satisfies the weak learning assumption)
that cannot be learned by the boosting algorithm.

In more detail, by \Cref{thm:gammaVCHS} there exists $N\subseteq \R^d$ of size $n:=\lvert N\rvert$ which
is nearly quadratic in $1/\gamma$ which is $\gamma$-interpolated by $d$-dimensional halfspaces:
that is, each of the $2^n$ labelings $c:N\to\{\pm 1\}$ are $\gamma$-realizable by $d$-dimensional halfspaces.
In other words, each of these $c$'s satisfy the weak learnability assumption with respect to a $\gamma$-learner for halfspaces. 
Therefore, given enough $c$-labelled examples,
our assumed boosting algorithm will generate a weighted majority of $T$ halfspaces $h$
which is $\eps$-close to it.

Let $\H_T$ denote the family of all functions $h:N\to\{\pm 1\}$ which can be represented by a weighted
majority of $T$ halfspaces. The desired bound on $T$ follows by upper and lower bounding the size of $\H_T$:
on the one hand, the above reasoning shows that $\H_T$ forms an $\eps$-cover of the family of all functions $c:N\to\{\pm 1\}$
in the sense that for every $c\in\{\pm 1\}^N$ there is $h\in \H_T$ that is $\eps$-close to it.
A simple calculation therefore shows $\H_T$ must be large (has at least some $\exp(n)$ functions).
On the other hand, we argue that the number of $h$'s that can be represented
by a (weighted) majority of $T$ halfspaces must be relatively small (as a function of $T$).
The desired bound on $T$ then follows by combining these upper and lower bounds.

We make two more comments about this proof which may be of interest.
\begin{itemize}
\item First, we note that the set $N$ used in the proof is a regular\footnote{Let us remark in passing that $N$ can be chosen more generally; the important property it needs to satisfy is that the ratio between the largest and smallest distance among a pair of distinct points in $N$ is $O(n^{1/d})$, see \cite[Chapter~6.4]{matousekDiscrepencyBook}.}
grid (this set is implied by \Cref{thm:gammaVCHS}).
Therefore, the hard learning tasks which require a large oracle complexity are natural: the target distribution is uniform over a regular grid.
\item The second comment concerns our upper bound on $\H_d$. 
Our argument here can be used to generalize a result by~\cite{Blumer89VC} regarding the composition of VC classes.
They showed that given classes $\B_1\ldots \B_T$ such that $\vc(\B_i)=d_i$ and a function $g:\{\pm 1\}^T\mapsto\{\pm 1\}$,
the class
\[\{g(b_1\ldots b_T) : b_i\in \B_i\}\]
has VC dimension $ O((d_1 + \ldots + d_T)\log T)$.
Our argument generalizes the above by allowing to replace $g$ by a class of functions $G=\{g:\{\pm 1\}^T\to\{\pm 1\}\}$ 
and showing that the class
\[\{g(b_1\ldots b_T) : b_i\in \B_i, g\in G\}\]
has VC dimension $O((d_1+\ldots + d_T + d)\log T)$, where $d=\vc(G)$. (See \Cref{prop:vccomp})
\end{itemize}

\subsubsection{Upper Bound}
\paragraph{\Cref{alg:sepBoost}.}

We next try to provide intuition for \Cref{alg:sepBoost} and discuss some technical aspects in its analysis.
The main idea behind the algorithm boils down to a simple observation:
	let $S=(x_1,y_1)\ldots (x_m,y_m)$ be the input sample.
	Let us say that $b_1\ldots b_T\in \B$ {\it separate} $S$
	if for every $x_i,x_j$ such that $y_i\neq y_j$ there exists $b_t$ such that $b_t(x_i)\neq b_t(x_j)$.
	That is, every pair of input examples that have opposite labels are separated by one of the weak hypotheses.
	The observation is that {\it $b_1\ldots b_T$ can be aggregated to an hypothesis $h=f(b_1\ldots b_T)$ which is consistent with $S$
	if and only if the $b_t$'s separate $S$}. 
	This observation is stated and proved in~\Cref{lem:sep}.

Thus, \Cref{alg:sepBoost} attempts to obtain as fast as possible weak hypotheses $b_1\ldots b_T$ that separate the input sample $S$.
	Once $S$ is separated, by the above observation the algorithm can find and return an hypothesis $h=f(b_{1},\ldots,b_{T})$ that
	is consistent with the input sample. 
	To describe \Cref{alg:sepBoost}, it is convenient to assign to the input sample $S$
	a graph $G=(V,E)$, where $V=[m]$ and $\{i,j\}\in E$ if and only if $y_{i}\neq y_{j}$. 
	The graph $G$ is used to define the distributions $P_t$ on which the weak learner is applied during the algorithm:
	at each round~$t$, \Cref{alg:sepBoost} feeds the weak learner with  a distribution $P_{t}$ over $S$, 
	where the probability of each example $(x_i,y_i)$ is proportional to the degree of $i$ in $G$. 
	After receiving the weak classifier $b_{t}\in\mathcal{B}$, the graph $G$ is updated
	by removing all edges $\{{i},{j}\}$ which are separated by $b_{t}$ (i.e., such that $b_t(x_i)\neq b_t(x_j)$). 
	This is repeated until no edges are left, at which point the input sample is separated by $b_t$'s and we are done. 
	The analysis of the number of rounds $T$ which are needed until all edges are separated 
	appears in \Cref{thm:mainOracleComp}. In particular it is shown that $T=O(\log m /\gamma)$ with high probability.

\paragraph{Generalization Guarantees.}
As noted earlier, \Cref{alg:sepBoost} is a meta-algorithm in the sense that it does not specify how to find the aggregation rule $f$ in Line \ref{line13}.
	In particular, this part of the algorithm may be implemented differently for different base-classes.
	We therefore provide generalization guarantees which adapt to the way this part is implemented.
	{\it In particular, we get better guarantees for simpler aggregation rules.}
	More formally, following \cite[Chapter~4.2.2]{Schapire2012} we assume that with every sequence of weak hypotheses $b_1\ldots b_T\in\B$
	one can assign an {\it aggregation class} 
	\[\H = \H(b_1,\ldots, b_T)\subseteq \Bigl\{ f(b_1\ldots b_T) : f:\{\pm 1\}^T\to \{\pm 1\}\Bigr\},\]
	such that the output hypothesis of \Cref{alg:sepBoost} is a member of $\H$.
	For example, in classical boosting algorithms such as Adaboost, $\H$ is the class of all weighted majorities
	$\{\sign\{\sum_i w_i\cdot b_i\} : w_i\in\R\}$. 
	Our aggregation-dependent generalization guarantee adapts to the capacity of $\H$: smaller $\H$ yield better guarantees.
	This is summarized in \Cref{prop:genalg}.
	From a technical perspective, the proof of \Cref{prop:genalg} hinges on the notion of hybrid-compression-schemes
	from \cite[Theorem~4.8]{Schapire2012}.
\vspace{2mm}

\begin{figure}
\centering
\includegraphics[width=0.9\textwidth]{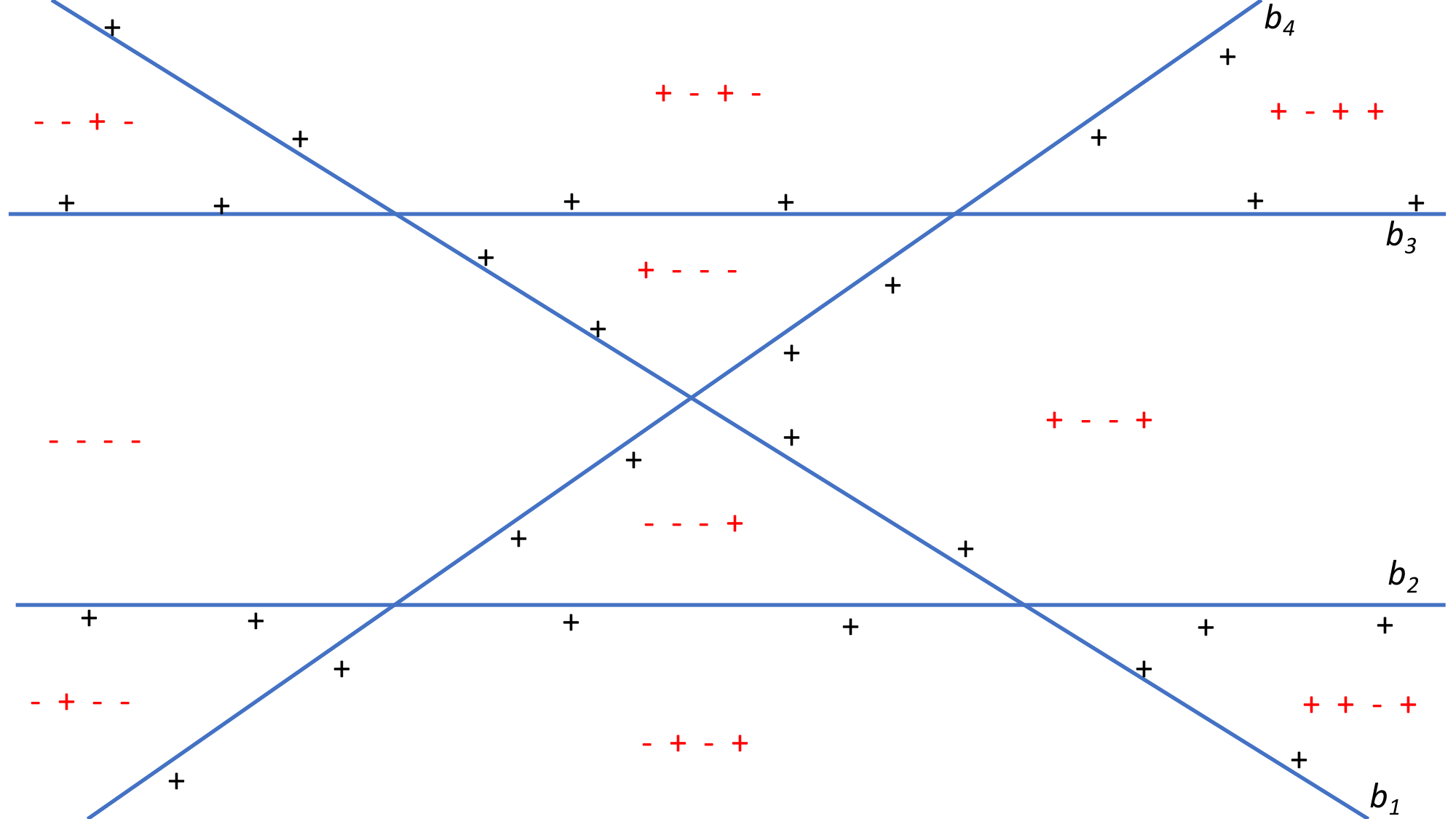}
\caption{\label{fig:cells}%
A set of 4 halfplanes $b_1\ldots b_4$ and the induced partition of $\R^2$ to cells, where $x',x''\in\R^2$ are in the same cell
if $\bigl(b_1(x'),b_2(x'),b_3(x'),b_4(x')\bigr)=\bigl(b_1(x''),b_2(x''),b_3(x''),b_4(x'')\bigr)$. Any hypothesis of the form $f(b_1, b_2, b_3, b_4)$ is constant on each cell in the partition.}
\end{figure}

Finally, we show that even without any additional restriction on $\B$ besides being a VC class,
	it is still possible to use \Cref{prop:genalg} to derive polynomial sample complexity.
	The idea here boils down to showing that given the weak hypotheses $b_1\ldots b_T\in\B$,
	one can encode any aggregated hypothesis of the form $f(b_1\ldots b_T)$ using its values on the {\it cells} defined by the $b_t$'s:
	indeed, the $b_t$'s partition $\X$ into cells, where $x',x''\in \X$ are in the same cell if and only if $b_t(x')=b_t(x'')$ for every $t\leq T$.
	For example, if the $b_t$'s are halfspaces in $\R^d$ then these are exactly the convex cells of the hyperplanes arrangement
	defined by the $b_t$'s. (See \Cref{fig:cells} for an illustration in the plane.) Now, since $\B$ is a VC class, one can show that the number of cells is at most $O(T^{d^*})$,
	where $d^*$ is the dual VC dimension of $\B$. This enables a description of any aggregation $f(b_1\ldots b_T)$
	using $O(T^{d^*})$ bits.\footnote{Note that $d^*=O(1)$ since $d^*< 2^{d+1}$ where $d=\vc(\B)=O(1)$, 
	and therefore the number of bits is polynomial in $T$~\citep{Assouad83densite}.
	We remark also that many natural classes, such as halfspaces, satisfy $d^* \approx d$.}
	The complete analysis of this part appears in \Cref{prop:vcagr,thm:mainGeneralization}. 

As discussed earlier, we consider that above bound of purely theoretical interest as it assumes that the aggregation rule is completely arbitrary.
	We expect that	for specific and structured base-classes $\B$ which arise in realistic scenarios,
	one could find consistent aggregation rules more systematically and get better generalization guarantees using \Cref{prop:genalg}.

\subsection{Expressivity}

We next overview some of main ideas which are used to analyze 
	the notions of $\gamma$-realizability and the $\gamma$-VC and $\gamma$-ID dimensions.

\paragraph{A Geometric Point of View.}
We start with a simple yet useful observation regarding the notion of $\gamma$-realizability:
	recall that a sample $S=((x_1,y_1)\ldots (x_m,y_m))$ is $\gamma$-realizable with respect to $\B$ if for every distribution $p$
	over $S$ there is an hypothesis $b\in \B$ which is $\gamma$-correlated with $S$ with respect to $p$.
	The observation is that this is \emph{equivalent} to saying that the vector $\gamma\cdot (y_1\ldots y_m)$
	(i.e., scaling $(y_1\ldots y_m)$ by a factor $\gamma$) belongs to the convex-hull of the set $\{(b(x_1)\ldots b(x_m)) : b\in \B\}$,
	i.e., it is a convex combination of the restrictions of hypotheses in $\B$ to the~$x_i$'s.
	This is proven by a simple Minmax argument in \Cref{lem:geom}.

This basic observation is later used to prove \Cref{prop:sufficientRank} via elementary linear algebra.
	(Recall that \Cref{prop:sufficientRank} asserts that under mild assumptions on $\mathcal{B}$,
	every sample $S$ is $\gamma$-realizable for a sufficiently small~$\gamma$.)
 
This geometric point of view is also useful in establishing the quadratic upper bound on the $\gamma$-VC dimension which is given in \Cref{thm:generalGammaVC}.
	The idea here is to use the fact that the scaled vector $\gamma\cdot (y_1\ldots y_m)$ can be written as a convex combination of the $b$'s
	to deduce (via a Chernoff and union bound) that $(y_1\ldots y_m)$ can be written as a majority vote of some $O(\log m/\gamma^2)$ of $b$'s in $\B$.
	Then, a short calculation which employs the Sauer--Shelah--Perles Lemma implies the desired bound.
	
\paragraph{Discrepancy Theory.}

There is an intimate relationship between Discrepancy theory and the $\gamma$-VC dimension:
	consider the problem of upper bounding the $\gamma$-VC dimension of a given class $\B$; 
	say we want to show that $\vc_\gamma(\B)< n$.
	In order to do so, we need to argue that for every $x_1\ldots x_n \in \X$ there are labels $y_1\ldots y_n\in \{\pm 1\}$
	such that the combined sample $S=(x_1,y_1)\ldots (x_n,y_n)$ is \emph{not} $\gamma$-realizable.
	That is, we need to show that $S$ exhibits $<\gamma$ correlation with \emph{every} $b\in \B$ with respect to \emph{some} distribution on $S$.	

How does this relate to Discrepancy theory?
	Let $F$ be a family of subsets over $[n]$, in the context of discrepancy theory, the goal is to assign a coloring $c:[n]\to \{\pm 1\}$
	under which every member $f\in F$ is balanced. 
	That is, for every $f\in F$ the sets $\{i\in f : c(i)=+1\}$ and $\{i\in f : c(i)=-1\}$ should be roughly of the same size.
	A simple argument shows that one can identify with every class $\B$ and $x_1\ldots x_n\in \X$ a family of subsets~$F$ over~$[n]$	such that a balanced coloring $c:[n]\to\{\pm1\}$ yields a sample $S=(x_1,c(1))\ldots (x_n,c(n))$ 
	which exhibits low correlation with every $b\in \B$ w.r.t.\ to the uniform distribution over $x_1\ldots x_n$.
	To summarize: 
	\begin{center}
	{\it  Balanced colorings imply upper bounds on the $\gamma$-VC dimension}.
	\end{center}
	A simple demonstration of this connection is used to prove \Cref{thm:ubdisc} which gives an
	upper bound on the $\gamma$-VC dimension with a subquadratic dependence on $\gamma$ (hence improving \Cref{thm:generalGammaVC}).

\vspace{2mm}

To conclude, the results in discrepancy are directly related to $\gamma$-realizability when the distribution over the sample $S$ is uniform.
	However, arbitrary distributions require a special care.
	In some cases, 	it is possible to modify arguments from discrepancy theory to apply to non-uniform distributions.
	One such example is our analysis of the $\gamma$-VC dimension of halfspaces in \Cref{thm:gammaVCHS}, 
	which is an adaptation of (the proof of) a seminal result in Discrepancy theory due to~\cite{Alexander90disc}.
	Other cases, such as the analysis of the $\gamma$-VC of decision stumps require a different approach.
	We discuss this in more detail in the next paragraph.

\paragraph{Linear Programming.}

\Cref{thm:gammaVCDS} provides a bound of $\Theta_d(1/\gamma)$ on the $\gamma$-VC dimension of the class $\ds_d$ of $d$-dimensional decision stumps 
	(i.e., axis aligned halfspaces).
	The upper bound (which is the more involved direction) is based on a geometric argument which may be interesting in its own right:
	let $m=\vc_\gamma(\ds_d)$; we need to show that if $A=\{x_1\ldots x_m\}\subseteq\R^d$ satisfies that each of the $2^m$ labelings of it are $\gamma$-realizable 
	by $\ds_d$ then $\gamma \leq O(d/m)$ (this implies that $m\leq O(d/\gamma)$ as required).
	In other words, we need to derive $m$ labels $\vec y = (y_1\ldots y_m)$ and a distribution $\vec p = (p_1\ldots p_m)$ over $\{x_1\ldots x_m\}$ 
	such that 
\begin{equation}\label{eq:1}
(\forall b\in \ds_d):\sum_i p_i\cdot y_i\cdot b(x_i) = O(d/m).
\end{equation}
	In a nutshell, the idea is to consider a small finite set of decision stumps $N\subseteq \ds_d$ of size $\lvert N\rvert \leq m/2$
	with the property that for every decision stump $b\in\ds_d$ there is a representative $r\in N$ such that the number of $x_i$'s
	where $b(x_i)\neq r(x_i)$ is sufficiently small (at most $O(m/d)$). That is, $b$ and $r$ agree on all but at most a $O(1/d)$ fraction of the $x_i$'s.
	The existence of such a set $N$ follows by a Haussler's Packing Lemma~\citep{Haussler95packing}.
	Now, since $\lvert N\rvert \leq m/2$, we can find many pairs $(\vec p, \vec y)$ such that
\begin{equation}\label{eq:2}
(\forall r\in N): \sum_i p_i\cdot y_i\cdot r(x_i) = 0.
\end{equation}
	This follows by a simple linear algebraic consideration (the intuition here is that there are only $m/2$ constraints in \Cref{eq:2} but $m$ degrees of freedom).
	We proceed by using a Linear Program to define a polytope which encodes the set of all pairs $(\vec p ,\vec y)$ which satisfy \Cref{eq:2},
	and arguing that a vertex of this polytope corresponds to a pair $(\vec p, \vec y)$ which satisfies \Cref{eq:1}, as required.

The above argument applies more generally for classes which can be represented as a small union of 1-dimensional classes (see \Cref{thm:gammaVC1}).
	
\section{Oracle-Complexity}\label{sec:oracle}

In this section we state and derive the oracle-complexity upper and lower bounds.
We begin with the upper bound in \Cref{sec:oracleub}, where we analyze \Cref{alg:sepBoost},
and then derive the lower bound in \Cref{sec:oraclelb}, where we also prove a combinatorial
result about composition of VC classes which may be of independent interest.

\subsection{Oracle Complexity Upper Bound}\label{sec:oracleub}
Our results on the expressivity of boosting advocate choosing a simple
base-class $\mathcal{B}$, and use it via boosting to learn concepts which may be far away
from $\B$ by  adjusting the advantage parameter $\gamma$. 
We have seen that the overall boosting algorithm becomes more expressive 
as $\gamma$ becomes smaller.
On the other hand, reducing $\gamma$ also increases the difficulty of weak learning: 
indeed, detecting a $\gamma$-correlated hypothesis in $\B$ amounts to solving an empirical risk minimization
problem over a sample of $O(\vc(\B)/\gamma^2)$ examples.
It is therefore desirable to minimize the number of times the weak learner is applied in the boosting procedure.

\paragraph{Improved Oracle Complexity Bound.}
The optimal oracle complexity was studied before in \cite[Chapter~13]{Schapire2012}, where it was shown that there exists a weak learner $\mathcal{W}$
such that the population loss of \emph{any} boosting algorithm after
$t$ interactions with $\mathcal{W}$ is at least $\exp(-O(t\gamma^{2}))$.

One of the main points we wish to argue in this manuscript is that 
one can ``bypass'' impossibility results by utilizing the simplicity of the weak hypotheses. 
We demonstrate this by presenting a boosting paradigm (\Cref{alg:sepBoost})
called "Graph-Separation Boosting" which circumvents the lower bound from \cite{Schapire2012}.

\setcounter{AlgoLine}{0}
\begin{algorithm}
\SetKwComment{Comment}{}{}
\SetKwInput{Parameters}{Parameters}
\Parameters{a base-class $\B$, a weak learner $\W$ with sample complexity $m_0$, an advantage parameter $\gamma>0$.}
\SetKwInput{WL}{Weak Learnability}
\WL{for every distribution $D$ which is $\gamma$-realizable by $\B$: $\E_{S'\sim D^{m_0}}\bigl[\mathsf{corr}_D\bigl(\W(S')\bigr)\bigr]\geq \gamma/2$.}
\KwIn{a sample $S=((x_{1},y_1),\ldots,(x_{m},y_m))$ which is $\gamma$-realizable by $\B$, and a black-box oracle access to the weak learner $\W$.}
Define an undirected graph $G=(V,E)$ where $V=[m]$ and $\{i,j\}\in E\Leftrightarrow y_{i}\neq y_{j}$.\;
Set $t\leftarrow0$.\; 
\While{$E\neq\emptyset$.}{ 
$t:=t+1$.\;
Define distribution $P_{t}$ on $S:~ P_{t}(x_i,y_i)\propto deg_G(i)$.\;
\tcp*[r]{($\deg_G(\cdot )$ is the degree in the graph $G$)}
Draw a sample $S_t \sim P_{t}^{m_0}$.\;
Set $b_{t}\leftarrow\mathcal{A}(S_t)$.\;
Remove from $E$ every edge $\{i,j\}$ such that $b_{t}(x_{i})\neq b_{t}(x_{j})$.
}
Set $T\leftarrow t$.\;
Compute an aggregation rule $f:\{\pm 1\}^T\to\{\pm 1\}$  such that the aggregated hypothesis $f(b_{1},\ldots b_{T})$ is consistent with $S$.\;
\tcp*[r]{($f$ exists by \Cref{{lem:sep}})}
Output $\hat{h}=f(b_{1},\ldots,b_{T})$.\;
\caption{Algorithm~\ref{alg:sepBoost} Restated}
\end{algorithm}

Similarly to previous boosting algorithms, the last step of our algorithm involves an aggregation of the hypotheses $b_1,\ldots,b_T$ returned by the weak learner $\mathcal{W}$ into a consistent classifier $h(x)=f(b_{1}(x),\ldots,b_{T}(x))$, where $f:\{\pm 1\}^T\to\{\pm 1\}$ is the aggregation function. While virtually all boosting algorithms (e.g., AdaBoost and
Boost-by-Majority) employ majority vote rules as aggregation functions, our boosting algorithm allows for more complex aggregation functions. 
This enables the quadratic improvement in the oracle complexity.

\vspace{2mm}

We now describe and analyze our edge separability-based boosting algorithm.
Throughout the rest of this section, fix a base-class $\mathcal{B}\subseteq\{\pm 1\}^{\mathcal{X}}$,
an edge parameter $\gamma >0$, and a weak learner denoted by~$\W$.
We let $m_0$ denote the sample complexity of $\W$ and assume that  for every distribution $D$ which is $\gamma$-realizable with respect to $\B$:
\begin{equation}\label{eq:9}
\E_{S\sim D^{m_0}}[\mathsf{corr}_D(\W(S))] \geq \gamma/2,
\end{equation}
where $\mathsf{corr}_D(h) = \E_{(x,y)\in D}[h(x)\cdot y]$ is the correlation of $h$ with respect to $D$.

The main idea behind the algorithm is simple. We wish to collect as
fast as possible a sequence of base classifiers $b_{1},\ldots,b_{T}\in\mathcal{B}$
that can be aggregated to produce a consistent hypothesis, i.e., a hypothesis
$h\in\{\pm1\}^{\X}$ satisfying $h(x_{i})=y_{i}$ for all $i\in[m]$.
The next definition and lemma provide a sufficient and necessary condition for
reaching such hypothesis. 
\begin{definition}
 Let $S=(x_{1},y_{1}),\ldots,(x_{m},y_{m})$
be a sample and let $b_{1},\ldots,b_{T}\in\{\pm 1\}^\X$ be hypotheses.
We say that $b_{1},\ldots,b_{T}$ separate $S$ if for every $i,j\in[m]$ with $y_{i}\neq y_{j}$,
there exists $t\in[T]$ such that~$b_{t}(x_{i})\neq b_{t}(x_{j})$. 
\end{definition}
\begin{lemma}[A Condition for Consistent Aggregation] \label{lem:sep}
Let $S=(x_{1},y_{1}),\ldots,(x_{m},y_{m})$ be a sample and let $b_{1},\ldots,b_{T}\in\{\pm 1\}^\X$ be hypotheses. 
Then, the following statement are equivalent.
\begin{enumerate}
\item There exists a function $h:=f(b_{1},\ldots,b_{T})\in\{\pm 1\}^{X}$ satisfying $h(x_{i})=y_{i}$ for every $i\in[m]$.
\item $b_{1},\ldots,b_{T}$ separate $S$.
\end{enumerate}
\end{lemma}
\begin{proof}
Assume that $b_1,\ldots,b_t$ separate $S$. 
Then, for any string $\bar{b} \in \{\pm 1\}^T$, the set \[
\{\,y_i\colon (b_1(x_i),\ldots,b_T(x_i))=\bar{b}\,\}
\] 
is either empty or a singleton. This allows us aggregating $b_1,\ldots,b_T$ into a consistent hypothesis. For example, we can define  
\[
f(\bar{b}) = \begin{cases} +1 & \exists i \in [m]~\textrm{s.t.}~(b_1(x_i),\ldots,b_T(x_i))=\bar{b} ~\& ~y_i = 1  \\ -1 & \textrm{otherwise} \end{cases}
\]
This proves the sufficiency of the separation condition. Suppose now that $b_1,\ldots,b_T$ do not separate $S$. This implies that there exist $i, j \in [m]$ such that $y_i \neq y_j$ and $(b_1(x_i),\ldots,b_T(x_i))=(b_1(x_j),\ldots,b_T(x_j))$. Then any classifier of the form $h =f(b_1,\ldots,b_T)$ must satisfy either $h(x_i) \neq y_i$ or $h(x_j)\neq y_j$.
\end{proof}

On a high level, \Cref{alg:sepBoost} attempts to obtain as fast as possible weak hypotheses $b_1\ldots b_T$ that separate the input sample $S=(x_1,y_1)\ldots (x_m,y_m)$.
To facilitate the description of \Cref{alg:sepBoost}, it is convenient to
introduce an undirected graph $G=(V,E)$, where $V=[m]$ and $\{i,j\}\in E$
if and only if $y_{i}\neq y_{j}$. 

The graph $G$ changes during the running of the algorithm:
on every round $t$, \Cref{alg:sepBoost} defines
a distribution $P_{t}$ over $S$, where the probability of each
example $(x_i,y_i)$ is proportional to the degree of $i$. 
Thereafter, the weak learner $\W$ is being applied on a subsample 
$S_t = (x_{i_1},y_{i_1})\ldots (x_{i_{m_0}}, y_{i_{m_0}})$ which is drawn i.i.d.\ according to $P_{t}$. 
After receiving the weak classifier $b_{t}\in\mathcal{B}$, the graph $G$ is updated
by removing all edges $\{{i},{j}\}$ such that $x_i,x_j$ are separated by $b_{t}$. 
This is repeated until no edges are left (i.e., all pairs are separated by some $b_t$). 
At this point, as implied by Lemma \ref{lem:sep}, \Cref{alg:sepBoost} 
can find and return an hypothesis $\hat{h}:=f(b_{1},\ldots,b_{T})\in\{\pm 1\}^{\X}$ that
is consistent with the entire sample.

\begin{theorem} [Oracle Complexity Upper Bound (\Cref{thm:mainOracleComp} restated)]  
Let $S$ be an input sample of size $m$ which is $\gamma$-realizable with respect to $\B$, 
and let $T$ denote the number of rounds \Cref{alg:sepBoost} performs when applied on~$S$. 
Then, for every $t\in\mathbb{N}$ 
\[\Pr[T\geq t] \leq \exp\bigl(2\log m -t \gamma/2\bigr).\]
In particular, this implies that $\E[T]=O(\log (m)/\gamma)$.
\end{theorem}
\begin{proof} 
Let $E_{t}$ denote the set of edges that remain in $G$ after the first $t-1$ rounds.
An edge $\{{i},{j}\}\in E_t$ is not removed on round $t$ only if $b_{t}$ errs either on $x_{i}$ or on $x_{j}$, namely 
\begin{equation}\label{eq:10}
\{i,j\}\in E_{t+1} \implies y_i\cdot b_t(x_i) + y_j\cdot b_t(x_j) \leq 0.
\end{equation}
Let $\mathsf{corr}_t(h) := \E_{x_i\sim P_t}[y_i\cdot h(x_i)]$. Therefore, by the definition of $P_t$:
\begin{align*}
\mathsf{corr}_{t}(b_t)=\sum_i P_t(x_i,y_i)b_t(x_i)y_i &= \frac{\sum_i deg_t(i)b_t(x_i)y_i}{\sum_i deg_t(i)} \tag{$deg_t(\cdot))$ denotes the degree in $E_t$.}\\
	     						&=\frac{\sum_{\{i,j\}\in E_t} \Bigl(b_t(x_i)y_i + b_t(x_j)y_j\Bigr)}{2\lvert E_t\rvert}\\
							&\leq\frac{2\lvert E_{t}\setminus E_{t+1}\rvert}{2\lvert E_t\rvert} = \frac{\lvert E_{t}\setminus E_{t+1}\rvert}{\lvert E_t\rvert}   \tag{by \Cref{eq:10}}
\end{align*}
Thus, $\mathsf{corr}_t(b_t) \leq \frac{\lvert  E_t\setminus E_{t+1} \rvert}{ \lvert E_t \rvert}$. 
Now, since $S$ is $\gamma$-realizable, \Cref{eq:9} implies that
\[ \E\Bigl[\mathsf{corr}_t(b_t) \Big\vert ~E_t\Bigr] \geq \frac{\gamma}{2}.\]
Therefore,
\begin{align*}  \E\Bigl[\frac{\lvert E_t\setminus E_{t+1}\rvert}{\lvert E_t\rvert} \Big\vert~ E_t\Bigr] \geq  \E\Bigl[\mathsf{corr}_t(b_t) \Big\vert~ E_t\Bigr] \geq \frac{\gamma}{2} 
&\implies 
\E\Bigl[\lvert E_t\setminus E_{t+1}\rvert \Big\vert ~ E_t\Bigr]\geq \frac{\gamma}{2}\cdot \lvert E_t\rvert\\
&\implies
\E\bigl[\lvert E_{t+1}\rvert \big\vert ~ E_t\bigr]\leq \Bigl(1-\frac{\gamma}{2}\Bigr)\cdot \lvert E_t\rvert
\end{align*}
Thus, after $t$ rounds, the expected number of edges
is at most $\binom{m}{2} \cdot (1-\gamma/2)^{t}$. 
Hence,  the total number of rounds~$T$ satisfies:
\[
\Pr[T \geq t] = 
\Pr[\lvert E_t\rvert > 0] \leq 
\E[\lvert E_t\rvert] \leq 
\binom{m}{2}\cdot \Bigl(1-\frac{\gamma}{2}\Bigr)^t 
\leq \exp\Bigl(2\log (m) - t \cdot \frac{\gamma}{2}\Bigr)\,,
\]
where in the second transition we used the basic fact that $\Pr[X>0] \leq \E[X]$ for every random variable $X\in\mathbb{N}$.
To get the bound on $\E[T]$, note that:
\[
\E[T] = 
\sum_{t=1}^{\infty}\Pr[T \geq t] \leq 
\sum_{t=1}^\infty \min\Bigl\{1, \binom{m}{2}\cdot (1-\gamma)^t \Bigr\} = 
O\Bigl(\frac{\log m}{\gamma}\Bigr),
\]
where in the first transition we used that $\E[X] = \sum_{t=1}^\infty \Pr[X\geq t]$ for every random variable $X\in\mathbb{N}$.
\end{proof}

\subsubsection{Aggregation-Dependent Generalization Bound}\label{sec:gen}
As  discussed in \Cref{sec:mainoracle}, \Cref{alg:sepBoost} is a meta-algorithm in the sense that it does not specify how to find the aggregation rule $f$ in Line \ref{line13}.
	In particular, this part of the algorithm may be implemented in different ways, depending on the choice of the base-class $\B$.
	We therefore provide here a generalization bound whose quality adapts to the complexity of this stage.
	That is, the guarantee given by the bound improves with the ``simplicity'' of the aggregation rule.

More formally, we follow the notation in \cite[Chapter~4.2.2]{Schapire2012} and assume that for every sequence of weak hypotheses $b_1\ldots b_T\in\B$
	there is an {\it aggregation class} 
	\[\H = \H(b_1,\ldots, b_T)\subseteq \Bigl\{ f(b_1\ldots b_T) : f:\{\pm 1\}^T\to \{\pm 1\}\Bigr\},\]
	such that the output hypothesis of \Cref{alg:sepBoost} is a member of $\H$.
	For example, for classical boosting algorithms such as Adaboost, $\H$ is the class of all weighted majorities
	$\{\sign(\sum_i w_i\cdot b_i) : w_i\in\R\}$. 

\begin{theorem}[Aggregation Dependent Bounds (\Cref{prop:genalg} restatement)]
Assume that the input sample $S$ to \Cref{alg:sepBoost} is drawn from a distribution $D$ which is $\gamma$-realizable with respect to $\B$.
Let $b_1\ldots b_T$ denote the hypotheses outputted by $\W$ during the execution of $\Cref{alg:sepBoost}$ on $S$,
and let $\H=\H(b_1\ldots b_T)$ denote the aggregation class.
Then, the following occurs with probability at least $1-\delta$:
\begin{enumerate}
\item{\bf Oracle Complexity}: the number of times the weak learner is called is
\[T=O\Bigl(\frac{\log m + \log(1/\delta)}{\gamma}\Bigr).\]
\item {\bf Sample Complexity}: The hypothesis $h\in\H$ outputted by \Cref{alg:sepBoost} satisfies $\mathsf{corr}_D(h) \geq 1-\eps$, where
\[
\eps = O \left (\frac{\bigl(T\cdot m_0 + \vc(\H)\bigr)\log m+\log(1/\delta)  }{m} \right) = \tilde O\Bigl(\frac{m_0 }{\gamma\cdot m} + \frac{\vc(\H)}{m}\Bigr),
\]
where $m_0$ is the sample complexity of the weak learner $\W$.
\end{enumerate}
\end{theorem}
\begin{proof}

Let $S\sim D^m$ be the input sample.
First, $S$ is $\gamma$-realizable and therefore by \Cref{thm:mainOracleComp}, the bound on $T$ in Item 1 holds with probability at least $1-\frac{\delta}{2}$.

For Item 2, we use the {\it hybrid-compression generalization bound} from \cite{Schapire2012}:
recall that in standard sample compressions, the output hypothesis is a function of a (short) tuple of the training examples.
Hybrid sample compression schemes are an extension of sample compression schemes 
in which the output hypothesis is instead selected from a class of hypotheses $\H$,
where the class (rather than the hypothesis itself) is a function of a (short) tuple of the training examples.
Specifically, we use the following result:
\begin{theorem}[{\cite[Theorem~4.8]{Schapire2012}}]
Suppose a learning algorithm based on a hybrid compression scheme of size $\kappa$ is provided with a random training set $S$ of size $m$. Suppose further that for every $\kappa$-tuple, the resulting class $\mathcal{F}$ has VC-dimension at most $d$.
Assume $m\geq d+ \kappa$. Then, with probability at least~$1-\beta$, any hypothesis $h$ produced by this algorithm that is consistent with $S$ has error at most
\[\frac{2d\log\bigl(2e(m-\kappa)/d\bigr) + 2\kappa\log m + 2\log(2/\beta)}{m-\kappa}.\]
\end{theorem}
To derive Item 2, notice that each~$b_i$ for $i=1,\ldots, T$ is determined by the tuple of the $m_0$ examples
which were fed as input to the weak learner $\W$ at the $i$'th iteration. Thus, the class $\H(b_1,\ldots b_T)$ is determined by the concatenated
tuple of $T\cdot m_0=:\kappa$ training examples.
Therefore, Theorem 4.8 in \cite{Schapire2012} implies\footnote{Note that in the bound stated in \Cref{prop:genalg} both $T$ and $\vc(\H)$ are random variables, while the corresponding parameters $\kappa$ and $d$ in Theorem~4.8 in \cite{Schapire2012} are fixed.
Thus, in order to apply this theorem, we use a union bound by setting $\delta_k= \frac{\delta}{100k^2}$, for each possible fixed value $k = T\cdot m_0 + \vc(\H)$.
The desired bound then follows simultaneously for all $k$ since $\sum_k \delta_k \leq \delta$.} that also the bound on $\eps$ in Item 2 holds with probability at least $1-\frac{\delta}{2}$. 
That is, with probability at least $1-\frac{\delta}{2}$:
\[\eps = O \left (\frac{\bigl(T\cdot m_0 + \vc(\H)\bigr)\log m+\log(1/\delta)  }{m} \right).\]
Thus, with probability at least $1-\delta$ both Items 1 and 2 are satisfied.
\end{proof}

\Cref{prop:genalg} demonstrates an upper bound on both the oracle and sample complexities of \Cref{alg:sepBoost}.
The sample complexity upper bound is algorithm-dependent in the sense that it depends on $\vc(\H)$
the VC dimension of $\H=\H(b_1\ldots b_T)$ -- the class of possible aggregations outputted by the algorithm.
In particular $\vc(\H)$ depends on the base-class $\B$ and on the implementation of Line \ref{line13} in \Cref{alg:sepBoost}.

One example where one can find a relatively simple aggregation class $\H$ is when $\B$ is the class of one-dimensional thresholds.
In this case, one can implement Line \ref{line13} such that $\vc(\H)=O(1/\gamma)$. 
This follows by showing that if $S$ is $\gamma$-realizable by thresholds then it has at most $O(1/\gamma)$ sign-changes and that one can choose $f=f(b_1\ldots b_T)$
to have at most $O(1/\gamma)$ sign-changes as well. So, $\H$ in this case is the class of all sign functions that change sign at most $O(1/\gamma)$ times
whose VC dimension is $O(1/\gamma)$.  
Note that in this example the bound on $\vc(\H)$ does not depend on $m$, 
which is different (and better) then the bound when $\H$ is defined with respect to aggregation by weighted majority.
More generally,  the following proposition provides a bound on $\vc(\H)$ 
when it is known that the aggregation rule belongs to a restricted class $G$:
\begin{proposition}[VC Dimension of Aggregation (\Cref{prop:vcagr} restatement)]
Let $\B \subseteq \{\pm 1\}^\X$ be a base-class and let~$G$ denote a class of ``$\{\pm 1\}^T\to\{\pm 1\}$'' functions (``aggregation-rules''). Then, 
\[\vc\Bigl(\Bigl\{g(b_1,\ldots,b_T) \vert b_i\in \B, g\in G\Bigr\}\Bigr) \leq  c_T\cdot(T\cdot \vc(\B) + \vc(G)),\]
where $c_T = O(\log T)$. 
Moreover, even if $G$ contains all ``$\{\pm 1\}^T\to \{\pm 1\}$'' functions,
	then the following bound holds for every fixed $b_1,b_2,\ldots, b_T\in \B$
\[
\vc\Bigl(\Bigl\{g(b_1,\ldots,b_T) \vert  g:\{\pm 1\}^T\to\{\pm 1\}\Bigr\}\Bigr) \le \binom{T}{\leq d^*} \leq 
(eT/d^*)^{d^*}\,,
\]
where $d^*$ is the dual VC dimension of $\B$.
\end{proposition}
\begin{proof}
The first part follows by plugging $\B_1=\B_2=\ldots=\B_T=\B$ in \Cref{prop:vccomp} which is stated in \Cref{sec:vccomp}.

For the second part, let $A\subseteq \X$ with 
$\lvert A\rvert>\binom{T}{\leq d^*}$. 
We need to show that $A$ is not shattered by the above class.
It suffices to show that there are distinct $x',x''\in A$ such that $b_i(x')=b_i(x'')$ for every $i\leq T$.
Indeed, by the Sauer--Shelah--Perles Lemma applied on the dual class of $\{b_1\ldots b_T\}$ we get that
\[ \Bigl\lvert\{(b_1(x),\ldots, b_T(x))  : x\in\X\} \Bigr\rvert \leq 
\binom{T}{\leq d^*} < \lvert A\rvert\,.
\]
Therefore, there must be distinct $x',x''\in A$ such that $b_i(x')=b_i(x'')$ for every $i\leq T$.
\end{proof}

The second part in \Cref{prop:vcagr} shows that even if the aggregation-rule used by \Cref{alg:sepBoost} is an arbitrary ``$\{\pm 1\}^T\to \{\pm 1\}$'' function,
one can still bound the VC dimension of all possible aggregations of any $T$ weak hypotheses $b_1\ldots b_T\in \B$
in terms of the dual VC dimension of $\B$ in a way that is sufficient to give generalization of \Cref{alg:sepBoost}
whenever $\B$ is a VC class. This is summarized in the following corollary.

\begin{corollary}[\Cref{thm:mainGeneralization} restatement]
Let $\B$ be the base-class, let $d^*$ denote its dual VC dimension, and assume oracle access to a $\gamma$-learner for $\B$
with sample complexity $m_0$.
Assume the input sample~$S$ to \Cref{alg:sepBoost} consists of $m$ examples drawn independently from a $\gamma$-realizable distribution.
Then with probability $1-\delta$ the following holds:
\begin{enumerate}
\item{\bf Oracle Complexity}: the number of times the weak learner is called is $T=O(\frac{\log m + \log(1/\delta)}{\gamma})$.
\item {\bf Sample Complexity}: The hypothesis $h\in\H$ outputted by \Cref{alg:sepBoost} satisfies $\mathsf{corr}_D(h) \geq 1-\eps$, where
\[
\eps = O \left (\frac{\bigl(T\cdot m_0 + T^{d^*}\bigr)\log m+\log(1/\delta)  }{m} \right) = \tilde O\Bigl(\frac{m_0 }{\gamma\cdot m} + \frac{1}{\gamma^{d^*}\cdot m}\Bigr),
\]
\end{enumerate}
\end{corollary}
As discussed earlier, we consider the above bound of purely theoretical interest as it assumes that the aggregation rule is completely arbitrary.
	We expect that	for specific and structured base-classes $\B$ which arise in realistic scenarios,
	one could find consistent aggregation rules more systematically and as a result to also get better guarantees on the capacity of the possible aggregation rules.

\subsection{Oracle Complexity Lower Bound}\label{sec:oraclelb}
We next prove a lower bound on the oracle complexity showing that if one restricts
	only to boosting algorithms which aggregate by weighted majorities then a near quadratic dependence in $1/\gamma$
	is necessary to get generalization, even if the base-class $\B$ is assumed to be a VC class.
	In fact, the theorem shows that even if one only wishes to achieve a constant error $\eps=1/4$
	with constant confidence $\delta=1/4$ then still nearly $1/\gamma^2$ calls to the weak learner are necessary,
	where $\gamma$ is the advantage parameter.

\begin{theorem}[Oracle Complexity Lower Bound (\Cref{thm:lowerOracleHalfspaces} restated)]
Let $\gamma>0$ be the edge parameter, and let $\B=\hs_d$ be the class of $d$-dimensional halfspaces.
	Let $\A$ be a boosting algorithm which uses a (possibly weighted) majority vote as an aggregation rule.
	That is, the output hypothesis of $\A$ is of the form
	\[h(x)=\sign\bigl( w_1\cdot b_1(x) + \cdots + w_T\cdot b_T(x) \bigr),\]
	where $b_1\ldots b_T$ are the weak hypotheses returned by the weak learner, and $w_1,\ldots w_T\in\mathbb{R}$.
	Then, for every weak learner $\W$ which outputs weak hypotheses from $\hs_d$ there exists a distribution $D$
	which is $\gamma$-realizable by $\hs_d$ such that if $\A$ is given sample access to $D$ and oracle access to $\W$, 
	then it must call $\W$ at least
	\[T=\tilde\Omega_d\Bigl(\frac{1}{\gamma^{2-\frac{2}{d+1}}}\Bigr)\]
	times in order to output an hypothesis $h$ such that with probability at least $1-\delta=3/4$ it satisfies $\mathsf{corr}_D(h)\geq 1-\eps=3/4$.
	The $\tilde \Omega_d$ above conceals multiplicative factors which depend on $d$ and logarithmic factors
	which depend on $1/\gamma$.
\end{theorem}

\begin{proof}
Let us strengthen the weak learner $\W$ by assuming that whenever it is given a sample from a $\gamma$-realizable distribution $D$
then it \emph{always} outputs a $h\in\hs_d$ such that $\mathsf{corr}_D(h)\geq \gamma$ (i.e., it outputs such an $h$ with probability~$1$).
Clearly, this does not affect generality in the context of proving oracle complexity lower bounds
(indeed, if the weak learner sometimes fails to return a $\gamma$-correlated hypothesis
then the number of oracle calls may only increase).

Let $T(\gamma,\eps,\delta)$ denote the minimum integer for which the following holds:
given sample access to a $\gamma$-realizable distribution $D$, 
the algorithm~$\A$ makes at most $T$ calls to $\W$ and outputs an hypothesis $h$ such that $\mathsf{corr}_D(h) \geq 1-\eps$ with probability at least $1-\delta$.
Thus, our goal is to show that $T=T(\gamma,1/4,1/4)\geq \tilde\Omega_d(1/\gamma^{\frac{2d}{d+1}})$.

By \Cref{thm:gammaVCHS} there exists $N\subseteq \R^d$ of size $n:=\lvert N\rvert = \Omega_d(1/\gamma^{\frac{2d}{d+1}})$
such that  each labeling $c:N\to\{\pm 1\}$ is $\gamma$-realizable by~$\hs_d$.
Let $u$ denote the uniform distribution over $N$.
Since for every $c:N\to\{\pm 1\}$ the distribution defined by the pair $(u,c)$ is $\gamma$-realizable 
it follows that given sample access to examples $(x,c(x))$ where $x\sim u$, 
the algorithm $\A$ makes at most $T$ calls to $\W$ and outputs~$h$ of the form
\begin{equation}\label{eq:11}
h(x)=\sign\bigl( w_1\cdot b_1(x) + \ldots + w_T\cdot b_T(x) \bigr) \quad \quad  b_i\in\hs_d,
\end{equation}
such that with probability at least $3/4$,
\[d(c,h):= \Pr_{x\sim u}[c(x)\neq h(x)] = \frac{1}{n}\bigl\lvert \bigl\{x\in  N: h(x) \neq c(x)\bigr\}\bigr\rvert \leq 1/4.\]
Let $\H_T$ denote the set of all functions $h:A\to\{\pm 1\}$ which can be represented like in \Cref{eq:11}.
The proof follows by upper and lower bounding the size of $\H_T$.
\paragraph{$\H_T$ is Large.}
By the above consideration it follows that
\[
(\forall c\in \{\pm 1\}^N)(\exists h\in \H_T) : d(c,h)\leq  1/4.
\]
In other words, each $c\in\{\pm 1\}^N$ belongs to a hamming ball of radius $1/4$ around some $h \in \H_T$. 
Thus, if $V(p)$ denotes the size of a hamming ball of radius $p$ in $\{\pm 1\}^N$, then $V(1/4)\cdot \lvert \H_T\rvert \geq 2^n$
and therefore
\begin{equation}\label{eq:12}
\lvert \H_T\rvert \geq  \frac{2^n}{V(1/4)} \geq2^{\bigl(1-\mathtt{h}_2(\frac{1}{4})\bigr)n},
\end{equation}
where $\mathtt{h}_2(x) = - x\log(x) - (1-x)\log(1-x)$ is the binary entropy function.
Indeed, \Cref{eq:12} follows from the basic inequality $V(p)\leq 2^{h(p)\cdot n}$ (see, e.g., \cite[Theorem 3.1]{Galvin14}).

\paragraph{$\H_T$ is Small.}
Let us now upper bound the size of $\H_T$:
each function in $\H_T$ is determined by
\begin{itemize}
\item[(i)] the restrictions to $N$ of the $d$-dimensional halfspaces $b_1\vert_N\ldots b_T\vert_N\in \{\pm 1\}^N$, and 
\item[(ii)] the $T$-dimensional halfspace defined by the $w_i$'s.
\end{itemize}
For (i), note that by the Sauer--Shelah--Perles Lemma, the total number of restriction of $b\in \hs_d$'s to $N$ is~$O(n^d)$ 
and therefore the number of ways to choose $T$ hypotheses $b_1\vert_N\ldots b_T\vert_N$ is $O(n^{d\cdot T})$.
For (ii), fix a sequence $b_1\vert_N\ldots b_T\vert_N$, and identify each $x\in N$ with the $T$-dimensional vector 
\[x\mapsto \bigl(b_i(x)\bigr)_{i=1}^{T}.\]
Thus, each function on the form
\[h(x)=\sign\bigl( w_1\cdot b_1(x) + \ldots + w_T\cdot b_T(x) \bigr) \]
corresponds in a one-to-one manner to a  halfspace in $T$-dimensions
restricted to the set 
\[B=\Bigl\{  \bigl(b_i(x)\bigr)_{i=1}^{T} : x\in N\Bigr\}\subseteq \R^T.\]
In particular, the number of such functions is $O(\lvert B\rvert^T) =O(\lvert N\rvert ^T) = O(n^T)$.
To conclude, 
\begin{equation}\label{eq:13}
\lvert \H_T\rvert \leq  O(n^{d\cdot T})\cdot O(n^T) = O(n^{(d+1)\cdot T}). 
\end{equation}
Combining \Cref{eq:12,eq:13} we get that
\[2^{n(1-\mathtt{h}_2(1/4))} \leq O(n^{(d+1)\cdot T}),\]
which implies that $T= \Omega(\frac{n}{d\log n}) = \tilde\Omega_d(1/\gamma^{\frac{2d}{d+1}})$
and finishes the proof.
\end{proof}

\subsubsection{The VC Dimension of Composition}\label{sec:vccomp}
We conclude this part by demonstrating how the argument used in the above lower bound can extend a classical
result by \cite{Blumer89VC}.

\begin{proposition}\label{prop:vccomp}
Let $\B_1\ldots \B_T \subseteq \{\pm 1\}^\X$ be classes of $\X\mapsto\{\pm 1\}$ functions and let
$G$ be a class of ``$\{\pm 1\}^T\to\{\pm 1\}$'' functions.
Then the composed class
\[G(\B_1\ldots \B_T)=\{g(b_1\ldots b_T) : b_i\in \B_i, g\in G\}\subseteq\{\pm1\}^\X\]
satisfies
\[\vc\bigl(G(\B_1\ldots \B_T)\bigr) \leq  c_T\cdot(\vc(\B_1) + \ldots + \vc(\B_T) + \vc(G)),\]
where $c_T = O(\log T)$.\footnote{Specifically, $c_T = \frac{1}{T\cdot x}$ where $x<1/2$ is such that $h(x)=\frac{1}{T+1}$,
and $h(\cdot )$ is the binary entropy function.}
\end{proposition}

This generalizes a result by \cite{Blumer89VC} who considered the case when $G=\{g\}$ consists of a single function.

\begin{proof}

Without loss of generality we may assume that each $d_i\geq 1$ (indeed, else $\lvert \B_i\rvert\leq 1$ and we may ignore it).
By the Sauer--Shelah--Perles Lemma, for every $A\subseteq \X$ and for every $i\leq T$
\[\lvert \B_i\vert_A\rvert \leq  
\binom{\lvert A\rvert}{\leq d_i} \leq 
2\lvert A\rvert^{d_i}.\] 
Similarly, for every $B\subseteq \{\pm 1\}^T$:
\[\lvert G\vert_B\rvert \leq \binom{\lvert B\rvert}{\leq{d_G}} \leq  2\lvert B\rvert^{d_G}.\] 

Let $N\subseteq \X$ of size $n:=\vc(G(\B_1\ldots \B_T))$. such that $N$ is shattered by $G(\B_1\ldots \B_T)$.
Thus,
\begin{equation}\label{eq:14}
\Bigl\lvert G(\B_1\ldots \B_T)\vert_N\Bigr\rvert = 2^n.
\end{equation}
On the other hand, note that each function $g(b_1\ldots b_T)\vert_N$ is determined by
\begin{itemize}
\item[(i)] the restrictions $b_1\vert_N\ldots b_T\vert_N\in \{\pm 1\}^N$, and 
\item[(ii)] the identity of the composing function $g\in G$ restricted to the set 
$\{(b_1\vert_N(x),\ldots, b_T\vert_N(x)) : x\in N\}\subseteq\{\pm 1\}^T$.
\end{itemize}
For (i), by the Sauer--Shelah--Perles Lemma the number of ways to choose $T$ restrictions $b_1\vert_N\ldots\allowbreak b_T\vert_N$ where~$b_i\in \B_i$ is at most
\[
\binom{n}{\leq d_1}\cdot 
\binom{n}{\leq  d_2}\cdot\cdots\cdot 
\binom{n}{\leq d_T} \,.
\]
For (ii), fix a sequence $b_1\vert_N\ldots b_T\vert_N$, and identify each $x\in N$ with the $T$-dimensional boolean vector 
\[x\mapsto \bigl(b_i(x)\bigr)_{i=1}^{T}\,.\]
By the Sauer--Shelah--Perles Lemma,
\[
\Bigl\lvert \Bigl\{g(b_1(x)\ldots b_T(x)) : g\in G, x\in N \Bigr\}\Bigr\rvert \leq \binom{n}{\leq d_G}\,.
\]
Thus, 
\begin{align}
\bigl\lvert G(\B_1\ldots\B_T)\vert_N \bigr\rvert &\leq  
\binom{n}{\leq d_1}\cdot\cdots\cdot \binom{n}{\leq d_T}
\cdot \binom{n}{\leq d_G}\nonumber \\
										&\leq 2^{n\cdot(h(d_1/n) +\ldots +h(d_T/n) + h(d_G/n))}\,, \label{eq:15}
\end{align}
where we used the basic inequality $\binom{n}{\leq k}\leq 2^{nh(k/n)}$, where $h(x)= -x\log x- (1-x)\log(1-x)$ is the entropy function.
Combining \Cref{eq:14,eq:15} we get:
\begin{align*}
1 &\leq h(d_1/n) + \cdots + h(d_T/n) + h(d_G/n)\\
   &\leq (T+1)\cdot h\Bigl(\frac{d_1 + \cdots + d_T + d_G}{T\cdot n}\Bigr)\,, \tag{by concavity of $h(\cdot)$}
\end{align*}
and therefore $n=\vc(G(\B_1\ldots \B_T))$ must satisfy $\frac{1}{T+1}\leq h\bigl(\frac{d_1 + \cdots + d_T + d_G}{T\cdot n}\bigr)$.
So, if we let $x < 1/2$ such that $h(x) = \frac{1}{T+1}$ then, since $h(\cdot)$ is monotone increasing on $(0,1/2)$, 
we have  $\frac{d_1 + \cdots + d_T + d_G}{T\cdot n}\geq x$.
Therefore, $n \leq  c_T\cdot (d_1 + \ldots + d_T + d_G)$, where $c_T = \frac{1}{T\cdot x} = O(\log T)$, as required.
\end{proof}

\section{Expressivity}\label{sec:expressivity}
Throughout this section we assume that the base-class $\B\subseteq\{\pm 1\}^\X$ is symmetric 
	in the following sense: 
	\[(\forall b\in\{\pm 1\}^\X): b\in \B \iff -b\in \B\,.\]
	Note that this assumption does not compromise generality because: (i)  a learning algorithm for $\B$
	implies a learning algorithm for $\{\pm b : b\in \B\}$,
	and (ii) $\vc(\{\pm b : b\in \B\})\leq \vc(\B) + 1$.
	So, if $\B$ is not symmetric, we can replace it by $\{\pm b : b\in \B\}$.
	
\paragraph{Organization.}
We begin with stating and proving a basic geometric characterization of $\gamma$-realizability in \Cref{sec:geom},
	which may also be interesting in its own right. This characterization is then used to prove \Cref{prop:sufficientRank}, 
	which implies that virtually all VC classes which are typically considered in the literature are expressive 
	when used as base-classes.
	Then, in \Cref{sec:genbounds} we provide general bounds on the growth rate of the $\gamma$-VC dimension.
	We conclude the section by analyzing the classes of Decision Stumps (\Cref{sec:ds})
	and of Halfspaces (\Cref{sec:hs}).

\subsection{A Geometric Perspective of \texorpdfstring{$\gamma$}{gamma}-realizability}\label{sec:geom}

The following simple lemma provides a geometric interpretation of $\gamma$-realizability and the $\gamma$-VC dimension,
	which will later be useful.

\begin{lemma}[A Geometric Interpretation of $\gamma$-Realizability]\label{lem:geom}
Let $\B\subseteq\{\pm 1\}^\X$ be a symmetric class and let $\gamma>0$.
\begin{enumerate}
\item A sample $S=((x_1,y_1)\ldots (x_n,y_n))$ is $\gamma$-realizable with respect to $\B$ 
if and only if there is a distribution~$q$ over $\B$ such that 
\[ (\forall i\leq n): \E_{b\sim q}[y_i\cdot b(x_i)] \geq \gamma.\]
Equivalently, $S$ is $\gamma$-realizable if and only if the vector $\gamma\cdot(y_1\ldots y_n) = (\gamma y_1\ldots \gamma y_n)$ 
is in the convex-hull of $\{(b(x_1)\ldots b(x_n)) : b\in \B\}$.
\item The $\gamma$-VC dimension of $\B$ is the maximum $d$ such that the continuous $\gamma$-cube $[-\gamma,+\gamma]^d$ satisfies 
\[[-\gamma,+\gamma]^d\subseteq \conv\Bigl(\bigl\{(b(x_1)\ldots b(x_d)) : b\in\B\bigr\}\Bigr)\] for some $x_1\ldots x_d\in \X$, where $\conv(\cdot)$ denote the convex hull operator.
\end{enumerate}
\end{lemma}

Note that this lemma can also be interpreted in terms of norms. 
	Indeed, since $\B$ is symmetric, the set
	\[ \conv\Bigl(\bigl\{(b(x_1)\ldots b(x_d)) : b\in\B\bigr\}\Bigr)\subseteq \R^d\]
	is a symmetric convex set and therefore defines a norm $\|\cdot\|$ on $\R^d$.
	Moreover, \Cref{lem:geom} implies that $(x_1,y_1)\ldots (x_d,y_d)$ is $\gamma$-realizable if and only if
	\[ \bigl\|(y_1\ldots y_d)\bigr\| \leq \frac{1}{\gamma}.\]
	Consequently, the $\gamma$-VC dimension of $\B$ is related to the {\it Banach-Mazur distance} (see e.g.\ \cite{Giannopoulos95compactum})
	of that norm from $\ell_\infty$ (e.g., if all samples $(y_1\ldots y_d)\in\{\pm 1\}^d$ are $\gamma$-realizable than that distance is at most $1/\gamma$).
	
\begin{proof}[Proof of Lemma\ref{lem:geom}]
The proof is a simple application of the Minmax Theorem~\citep{Neumann1928}:
for a sample $S=((x_1,y_1)\ldots (x_n,y_n))$ define a zero-sum two-player game,
where player 1 picks $b\in \B$ and player 2 picks $i\leq n$, and player's 2 loss is $y_i\cdot b(x_i)$.
Notice that $\gamma$-realizability of $S$ amounts to 
\[\min_{p\in \Delta_n} \max_{b\in \B}\E_{i\sim p}[y_i\cdot b(x_i)] \geq \gamma,\]
where $\Delta_n$ denotes the $n$-dimensional probability simplex.
By the Minmax Theorem, the latter is equivalent to
\[\max_{q\in \Delta(\B)} \min_{i\in [n]}\E_{b\sim q}[y_i\cdot b(x_i)] \geq \gamma,\]
where $\Delta(\B)$ is the family of distributions over $\B$.
Thus, $S$ is $\gamma$-realizable if and only if there is a distribution~$q$ over $\B$
such that $\E_{b\sim q}[y_i\cdot b(x_i)] \geq \gamma$ for every $i\leq n$.
Since $\B$ is symmetric, the latter is equivalent to the existence of $q'$
such that $\E_{b\sim q'}[y_i\cdot b(x_i)] = \gamma$ for every $i\leq n$.
This finishes the proof of the Item 1.
Item 2 follows by applying Item 1 on each of the $2^d$ vectors $(y_1\ldots y_d)\in \{\pm 1\}^d$.
\end{proof}

\subsubsection{A Condition for Universal Expressivity}

The following proposition asserts that under mild assumptions on $\mathcal{B}$,
    every sample $S$ is $\gamma$-realizable for a sufficiently small $\gamma=\gamma(S)>0$.
    This implies that in the limit as $\gamma\to 0$, 
    it is possible to approximate any concept using weak-hypotheses from $\mathcal{B}$.\footnote{More precisely, it is possible to interpolate arbitrarily large finite restriction of any concept. We note in passing that a result due to \cite{Bartlett07consistent} provides an infinite version of the same phenomena: under mild assumptions on the base-class $\mathcal{B}$, they show that a variant of AdaBoost is universally consistent.}

\begin{proposition}[A Condition for Universality (\Cref{prop:sufficientRank} restatement)]
The following statements are equivalent for a symmetric class~$\mathcal{B}$:
\begin{enumerate}
    \item For every $c:X\to\{\pm 1\}$ and every sample $S$ labelled by $c$, there is $\gamma>0$
    such that~$S$ is $\gamma$-realizable by $\mathcal{B}$.
    \item For every $\{x_1,\ldots,x_n\} \subseteq \mathcal{X}$, the linear-span of $\{(b(x_1),\ldots,b(x_n))\in\mathbb{R}^n:~b\in \mathcal{B}\}$ is $n$-dimensional.
\end{enumerate}
\end{proposition}
Observe that every class $\mathcal{B}$ that contains singletons or one-dimensional thresholds satisfies Item 2
    and hence also Item 1. Thus, virtually all standard hypothesis classes that are considered in the literature satisfy it.

\begin{proof} 
We begin with the direction $1\implies 2$. Let $\{x_1\ldots x_n\}\subseteq \X$.
By assumption, for every $(y_1\ldots y_n)\in \{\pm 1\}^n$ there is $\gamma>0$ such that
the sample $((x_1,y_1)\ldots (x_n,y_n))$ is $\gamma$-realizable.
Thus, by $\Cref{lem:geom}$, Item 1 there are coefficients $\alpha_b \geq 0$ for $b\in \B$
such that $\sum_{b\in B} \alpha_b\cdot b(x_i) = y_i$ for every $i$.
This implies that every vector $(y_1\ldots y_n)\in\{\pm 1\}^n$ is in the space spanned by
$\{(b(x_1),\ldots,b(x_n))\in\mathbb{R}^n:~b\in \mathcal{B}\}$
and hence this space is $n$-dimensional as required.

We next prove $2\implies 1$: let $S=((x_1,c(x_1)\ldots (x_n,c(x_n)))$ be a sample labeled by a concept $c$.
	We wish to show that $S$ is $\gamma$-realizable for some $\gamma>0$.
	By assumption, the set $\{(b(x_1)\ldots b(x_n)) : b\in B\}$ contains a basis,
	and hence there are coefficients $\alpha_b\in \R$
	such that 
	\[\sum{\alpha_b\cdot b(x_i) = c(x_i)}\] for every $i\leq n$.
	By possibly replacing $b$ with $-b$, we may assume that
	the coefficients $\alpha_b$ are nonnegative.
	By dividing $\alpha_b$ by $\sum_{b\in \B}\alpha_b$
	it follows that the vector 
	\[\frac{1}{\sum_{b\in \B}\alpha_b}\cdot (c(x_1)\ldots c(x_n))\]
	is in the convex hull of $\{(b(x_1)\ldots b(x_n)) : b\in \B\}$,
	which by \Cref{lem:geom} implies that $S$ is $\gamma$-realizable
	for~$\gamma=\frac{1}{\sum_{b\in \B}\alpha_b}$.
\end{proof}

\subsection{General Bounds on the \texorpdfstring{$\gamma$}{gamma}-VC Dimension}\label{sec:genbounds}

In the remainder of this section we provide bounds on the $\gamma$-VC dimension
    for general as well as for specific well-studied classes.
    As we focus on the dependence on~$\gamma$, we consider the VC dimension $d$ to be constant.
    In particular, we will sometimes use asymptotic notations~$O_d, \Omega_d$ which conceal multiplicative factors that depend on~$d$.

\begin{theorem}
[\Cref{thm:generalGammaVC} restatement]
Let $\mathcal{B}$ be a class with VC dimension $d$.
    Then, for every~$0<\gamma\leq 1$:
    \[\mathrm{VC}_{\gamma}(\mathcal{\mathcal{B}})=O\left(\frac{d}{\gamma^{2}}\log(d/\text{\ensuremath{\gamma}})\right) = \tilde O\Bigl(\frac{d}{\gamma^2}\Bigr).\]
    Moreover, this bound is nearly tight as long as $d$ is not very small comparing to $\log(1/\gamma)$: 
    for every $\gamma>0$ and $s\in\mathbb{N}$ there is a class $\mathcal{B}$ of VC dimension $d = O(s\log(1/\gamma))$
    and 
    \[\mathrm{VC}_{\gamma}(\mathcal{B})=\Omega\left(\frac{s}{\gamma^{2}}\right) = \tilde \Omega\Bigl(\frac{d}{\gamma^2}\Bigr).\]
\end{theorem}
Thus, the fastest possible growth of the $\gamma$-VC dimension is asymptotically $\approx d/\gamma^2$.
	We stress however that the above lower bound is realized by a class $\B$ whose VC dimension is at least $\Omega(\log(1/\gamma))$,
	which deviates from our focus on the setting the VC dimension is a constant and $\gamma\to 0$.
	Thus, we prove the next theorem which provides a sharp, subquadratic, dependence on~$\gamma$ (but a looser dependence on $d$). 
\begin{theorem}[$\gamma$-VC dimension: improved bound for small $\gamma$ (\Cref{thm:ubdisc} restatement)]
Let $\mathcal{B}$ be a class with VC dimension $d\ge 1$. Then, for every~$0<\gamma\leq 1$:
\[
\mathrm{VC}_{\gamma}(\mathcal{B}) \leq O_d\left( \left(\frac{1}{\gamma}\right)^{\frac{2d}{d+1}} \right),
\]
where $O_d(\cdot)$ conceals a multiplicative constant that depends only on $d$.
Moreover, the above inequality applies for any class $\B$ whose primal shatter function\footnote{The {\it primal shatter function} of a class $\B\subseteq\{\pm 1\}^\X$
is the minimum $k$ for which there exists a constant $C$ such that for every finite $A\subseteq \X$, the size of $\B|_A = \{b|_A : b\in \B\}$ is at most $C\cdot \lvert A\rvert^k$.
Note that by the Sauer-Shelah-Perles Lemma,  the primal shatter function is at most the VC dimension.} is at most $d$.
\end{theorem}
As follows from \Cref{thm:gammaVCHS}, the dependence on $\gamma$ in the above bound is tight.

\subsubsection{Proof of Theorem~\ref{thm:generalGammaVC}}
To prove the  upper bound, let $\B$ have VC dimension $d$, let $\gamma>0$, and let $I \subseteq \X$  
	be a set of size $\vc_\gamma(\B)$ such that every labeling of it is $\gamma$-realizable by $\B$.
	Fix $c: I \to \{\pm 1\}$. By \Cref{lem:geom} there is a probability distribution~$q$ on $\B$ so that 
	\[(\forall x\in I): \E_{b\sim q}\bigl[b(x)\cdot c(x)\bigr] \geq \gamma.\]
	This implies, using a Chernoff and union bounds, that $c$ is a majority of $O(\frac{\log \lvert I \rvert}{\gamma^2})$ restrictions of hypotheses in~$\B$ to~$I$. 
	As this holds for any fixed $c$ it follows that each of the $2^{\lvert I\rvert}$ distinct $\pm 1$ patterns on $I$ is
	the majority  of a set of at most $O(\frac{\log \lvert I\rvert}{\gamma^2})$ restrictions of hypotheses in $\B$ to $I$. 
	By the Sauer-Perles-Shelah Lemma \citep{Sauer72Lemma} there are less than $(e\lvert I\rvert/d)^d$ such restrictions, and hence
	\[ \Bigl[\Bigl(\frac{e\lvert I\rvert }{d}\Bigr)^d\Bigr]^{O(\log \lvert I\rvert /\gamma^2)} \geq 2^{\lvert I\rvert}.\]
	This implies that 
	\[ \lvert I\rvert \leq O\Bigl(\frac{d}{\gamma^2} \log \bigl(\frac{d}{\gamma^2}\bigr)\Bigr),\]
	completing the proof of the upper bound.

\vspace{2mm}

To prove the lower bound we need the following simple lemma.
\begin{lemma}
\label{l21}
Let $v_1,v_2, \ldots v_t$ be pairwise orthogonal vectors
in $\{\pm 1\}^t$. Then for every probability distribution
$p=(p_1,p_2, \ldots ,p_t)$ there is an $i$ so that
the absolute value of the inner product of $v_i $ and $p$ is at least
$\frac{1}{\sqrt t}$.
\end{lemma}
Note that such vectors exist if and only if there is a $t\times t$ Hadamard matrix.
In particular they exist for every $t$ which is a power of $2$ (and 
conjectured to exist for all $t$ divisible by $4$).
\begin{proof}
 Since the vectors $v_i/\sqrt t$ form an orthonormal basis,
\[
\sum_{i=1}^t \frac{1}{t} (v_i,p)^2 =\|p\|_2^2 =\sum_{i=1}^t p_i^2
\geq \frac{(\sum_{i=1}^t p_i)^2}{t}=\frac{1}{t}.
\]
Thus, there is an $i$ so that the inner product $\langle v_i,p\rangle^2 \geq \frac{1}{t}$, as needed.
\end{proof}

\begin{corollary}
\label{c22}
If $t=1/\gamma^2$ is a power of $2$ then there is a collection of
$2t$ vectors $u_i$ with $\{\pm 1\}$-coordinates, each
of length $t$ so that for every vector $h \in \{\pm 1\}^t$
and every probability distribution $p=(p_1,p_2, \ldots ,p_t)$
on its coordinates there is a vector $u_i$ so that 
$\E_{j\sim p}[h(j)\cdot v_i(j) ]  \geq \gamma.$
\end{corollary}
\begin{proof}
Let $v_1, \ldots ,v_t$ be the rows of a $t\times t$
Hadamard matrix, and consider the $2t$ vectors in the set $\{\pm v_i : i\leq t\}$.
The desired result follows from \Cref{l21}.
\end{proof}

We can now prove the lower bound in \Cref{thm:generalGammaVC}. 
\begin{proof}[Proof of Theorem~\ref{thm:generalGammaVC}]
	Let $t=1/\gamma^2$ be a power of $2$, let $s\in\mathbb{N}$, and put $m=s\cdot t$. 
	Fix a set $F$ of $2t$ vectors of length $t$ satisfying the assertion of
	\Cref{c22} and let $\B$ be the collection of all vectors obtained by concatenating $s$ members of $F$ (thus $\lvert \B\rvert =(2t)^s$).  
	By applying the above corollary to each of the~$s$ blocks of $t$ consecutive indices it is not difficult to check that
	for every vector $c \in \{\pm 1\}^m$  and for any probability
	distribution $p=(p_1, \ldots ,p_m)$, there is $b \in \B$  so that $\E_{i\sim p}[b_i\cdot c_i] \geq \gamma$.
	Therefore, we conclude that:
\begin{align*}
\vc(\B) &\leq \log \lvert \B\rvert = s\log\frac{2}{\gamma^2}, \\
\vc_\gamma(\B) &\geq s\cdot t =s \cdot \frac{1}{\gamma^2} \geq \frac{\vc(\B)}{\gamma^2 \log (2/\gamma^2)}.
\end{align*}
	This completes the  proof of \Cref{thm:generalGammaVC}.  
\end{proof}

\subsubsection{Proof of Section~\ref{thm:ubdisc}: An Improved Bound using Discrepancy Theory}\label{sec:discub}
There is an intimate relationship between the $\gamma$-VC dimension and {\it Discrepancy Thoery}
	(see, e.g., the book \cite{matousekDiscrepencyBook}). As a first application of this relationship, we prove \Cref{thm:ubdisc} by a simple reduction	
	to a classical result in Discrepancy Theory.
	We begin by introducing some notation.
	Let $F$ be a family of sets over a domain $A$ and let $n$ denote the size of $A$. 
	Discrepancy theory studies how {\it balanced} can a coloring of~$A$ be with respect to $F$.
	That is, for a coloring $c:A\to \{\pm 1\}$ and a set $f\in F$ define the discrepancy of $c$ with respect to~$f$ by
	\[\disc(c;f) = \Bigl\lvert\sum_{x\in f} c(x) \Bigr\rvert.\]
	Define the discrepancy of $c$ with respect to $F$ by
	\[\disc(c;F) = \max_{f\in F}\disc(c;f).\]
	Finally, the discrepancy of $F$ is defined as the discrepancy of the ``best'' possible coloring:
	\[\disc(F) = \min_{c:A\to\{\pm1\}} \disc(c;F).\]
	
\paragraph{Low Discrepancy implies large $\gamma$-VC Dimension.}
A classical result due to \cite{MWW93disc,Matousek95discrepancy}
asserts that every family $F$ of subsets over $A$ with a small VC dimension admits a relatively balanced coloring:
\begin{equation}\label{eq:3}
\disc(F) \leq C_d \cdot \lvert A\rvert^{\frac{1}{2} - \frac{1}{2d}},
\end{equation}
where $d$ is the VC dimension of $A$ and $C_d$ is a constant depending only on $d$ (see also Theorem 5.3 in \cite{matousekDiscrepencyBook}).
Let $\B\subseteq\{\pm 1\}^\X$ be a (symmetric) class and let $d:=\vc(\B)$. 
Let $A\subseteq \X$ be a set of size $\lvert A\rvert = \vc_\gamma(\B)$ such that each
of the $2^{\lvert A\rvert}$ possible labelings of $A$ are $\gamma$-realizable by $\B$.
Pick a coloring $c:A\to\{\pm 1\}$ which witnesses \Cref{eq:3} with respect to the family
\[F:= \bigl\{\supp(b) : b\in \B\bigr\},~\text{ where } \supp(b) = \{x\in A : b(x)=1\}.\]
Note that since $\B$ is symmetric, it follows that $\supp(b),\supp(-b)\in F$ for every $b\in\B$,
and also note that~$\vc(F)=\vc(\B)=d$.
Let $p$ denote the uniform distribution over $A$. For every $b\in \B$:
\begin{align*}
\E_{x\sim p}[c(x)\cdot b(x)] &=\frac{1}{\lvert A\rvert}\sum_{x\in A}b(x)c(x)\\
				&=\frac{1}{\lvert A\rvert}\sum_{x\in A: b(x)=1}c(x)-\frac{1}{\lvert A\rvert}\sum_{x\in A: b(x)=-1}c(x)\\
				&\leq\frac{1}{\lvert A\rvert}\disc(c;\supp(b)) + \frac{1}{\lvert A\rvert}\disc(c;\supp(-b))\\
				&\leq \frac{1}{\lvert A\rvert}\cdot 2C_d\lvert A\rvert^{\frac{1}{2} - \frac{1}{2d}} = 2C_d \lvert A\rvert^{-\frac{1}{2} - \frac{1}{2d}} \tag{by~\Cref{eq:3} applied on the family $F$.}.
\end{align*}
In particular, as by assumption, the sample $(x,c(x))_{x\in A}$ is $\gamma$-realizable, 
it follows that $\gamma \leq 2C_d \lvert A\rvert^{-\frac{1}{2} + \frac{1}{2d}}$ and therefore
\[\vc_\gamma(\B)=\lvert A\rvert \leq O_d\left( \left(\frac{1}{\gamma}\right)^{\frac{2d}{d+1}} \right)\]
as required.
 $\Box$

\subsection{Decision Stumps}\label{sec:ds}

We next consider the class of {\it Decision Stumps}.
A $d$-dimensional decision stump is a concept of the form $\sign(s(x_{j}-t))$, where $j\leq d$, $s\in\{\pm 1\}$ and $t\in\R$.
In other words, a decision stump is a halfspace which is aligned with one of the principal axes.
This class is popular in the context of boosting, partially because it is easy to learn it, even in the agnostic setting.
Also note that the Viola-Jones framework hinges on a variant of decision stumps~\citep{Viola01rapidobject}.
\begin{theorem}[\Cref{thm:gammaVCDS} restatement]
Let $\mathsf{DS}_d$ denote the class of decision stumps in $\mathbb{R}^d$ and $\gamma\in(0,1]$. 
Then, 
\[\mathrm{VC}_{\gamma}(\mathsf{DS}_d)=O\left( \frac{d}{\gamma}\right).\]
Moreover, the dependence on $\gamma$ is tight, already in the 1-dimensional case.
In fact, for every~$\gamma$ such that $1/\gamma\in \mathbb{N}$
\[\mathsf{ID}_\gamma(\ds_1)\geq  1/\gamma.\] 
For $d>1$, the class of $d$-dimensional decision-stumps $\gamma$-interpolates every set $Y\subseteq \mathbb{R}^d$ of size $1/\gamma$, 
provided that there exists $i\leq d$ so that every pair of distinct points $x,y\in Y$ satisfy $x_i\neq y_i$.
\end{theorem}
The proof of \Cref{thm:gammaVCDS} follows from a more general result concerning the union of classes with  VC dimension equal to~1. 
	We note that the bounds are rather loose in terms of $d$: 
	the upper bound yields a bound of $O(d/\gamma)$ while the lower bound gives only $\Omega(1/\gamma)$. 
	Also note that since the VC dimension of decision stumps is $O(\log d)$ (see \cite{Gey18stumps} for a tight bound), 
	\Cref{thm:generalGammaVC} implies an upper bound of $\tilde O(\log d /\gamma^2)$.
	It would be interesting to tighten these bounds.

\subsubsection{Proof of Theorem~\ref{thm:gammaVCDS}}

\paragraph{Lower Bound on $\mathsf{ID}_\gamma(\ds_1)$.}
We need to show that for every $\gamma$ such that $1/\gamma\in\mathbb{N}$ every set $A\subseteq \R$ of size $1/\gamma$
satisfies that each of the $2^{\lvert A\rvert}$ labeling of $A$ are $\gamma$-realizable by 1-dimensional decision stumps (i.e., thresholds).
Indeed, let $A=\{x_1<\ldots <x_m\}\subseteq \R$, let $(y_1\ldots y_m)\in\{\pm 1\}^m$, and let $p=(p_1\ldots p_m)$ be a distribution on $A$.
We need to show that there exists a threshold $b\in \ds_1$ such that $\E_{x_j\sim p}[y_j\cdot b(x_j)]\geq 1/m$.
Consider the $m+1$ sums
\[S_i=\sum_{j=1}^{i} y_j\cdot p_j  - \sum_{j=i+1}^{m} y_j\cdot p_j,  \qquad 0\leq i\leq m,\]
Note that since $\max_i \lvert S_i - S_{i-1}\rvert =\max_i 2p_i \geq 2/m$, there must be $i$ such that $\lvert S_i\rvert \geq 1/m$.
The proof of the lower bound is finished by noting that $\lvert S_i\rvert = \E_{x_j\sim p}[y_i\cdot b_i(x_j)]$, where 
\[
b_i(x)=
\begin{cases}
\sign \bigl(x - \frac{x_i+x_{i+1}}{2}\bigr) &S_j >0,\\
\sign\bigl(-(x-\frac{x_i+x_{i+1}}{2})\bigr) &S_j < 0.
\end{cases}
\]

The case of $d>1$ follows by a simple reduction to the $d=1$ case:
let $Y\subseteq\mathbb{R}^d$ be of size $1/\gamma$ such that there exists $i\leq d$ for which every pair of distinct points $x,y\in Y$ satisfy $x_i\neq y_i$.
Then, by projecting $Y$ on the first coordinate we obtain a $1$-dimensional set which is $\gamma$-interpolated by $\ds_1$ by the above argument.
Equivalently, $Y$ is $\gamma$-interpolated by decision-stumps that are aligned with the $i$th axis.
 $\Box$

\paragraph{Upper Bound on $\vc_\gamma(\ds_d)$.}
The upper bound is a corollary of the next proposition:
\begin{proposition}
\label{thm:gammaVC1}  Let $\mathcal{B}=\bigcup_{i=1}^{d}\mathcal{B}_{i}$
where for all $i\in[d]$, $\mathrm{VC}(\mathcal{B}_{i})\le1$. Then
$\mathrm{VC}_{\gamma}(\mathcal{B})\le O(d/\gamma).$ 
\end{proposition}

Note that \Cref{thm:gammaVC1} implies the upper bound \Cref{thm:gammaVCDS} since 
\[\ds_d = \bigcup_{j=1}^d \Bigl(\bigl\{\sign(x_j - t) : t\in \R\bigr\} \cup \bigl\{\sign(-(x_j - t)) : t\in \R\bigr\} \Bigr),\]
and each of the $2d$ classes that participate in the union on the right-hand side has VC dimension~$1$.

The proof of \Cref{thm:gammaVC1} uses Haussler's Packing Lemma, which we recall next.
	Let $p$ be a distribution over $\mathcal{X}$. 
	$p$ induces a (pseudo)-metric over $\{\pm 1\}^\X$, where the distance between $b',b''\in\{\pm 1\}^\X$ is given by
	\[d_p(b',b'')=p\bigl(\{x:~b'(x)\neq b''(x)\}\bigr).\] 
\begin{lemma}[Haussler's Packing Lemma~\citep{Haussler95packing}]\label{lem:packing}
Let $\mathcal{B}$ be a class of VC dimension $d$ and let $p$ be a distribution on $\mathcal{X}$.
Then, for any $\epsilon>0$ there exists a set $N=N(\eps,p)\subseteq \B$ of size $\lvert N\rvert \leq (20/\eps)^d$
such that
\[(\forall b\in \B)(\exists r\in N) : d_p(b,r)\leq \eps.\]
Such a set $C$ is called an $\eps$-cover for $\B$ with respect to $p$.
\end{lemma}

\begin{proof}[Proof of Theorem~\ref{thm:gammaVC1}] 
Let $A=\{x_{1},\ldots,x_{m}\}\subseteq\mathcal{X}$ be a set of size $m:=\vc_\gamma(\B)$ such that each of the $2^m$ possible labelings of it are $\gamma$-realizable by $\B=\cup_{i\leq d} \B_i$. We need to show that $\gamma \leq O(d/m)$.
By applying \Cref{lem:packing} with respect to the uniform distribution over $A$,
we conclude that for every class $\B_j$ there is $N_j\subseteq \B_j$ such that $\lvert N_j\rvert \leq \frac{m}{2d}$, and
\[
(\forall b\in\mathcal{B}_{j})(\exists r\in N_{j}):\frac{1}{m}\,\Bigl\lvert\bigl\{i:~b(x_{i})\neq r(x_{i})\bigr\}\Bigr\rvert\le\frac{20}{m/2d} = \frac{40d}{m}.
\]
The proof idea is to derive labels $(y_1\ldots y_m)\in\{\pm 1\}^m$ and a distribution $p$ over $A$ 
	such that (i) for every $j$, every $r\in N_j$ satisfies $\E_{x_i\sim p}[y_i\cdot r(x_i)]=0$, 
	and (ii) $p$ is sufficiently close to the uniform distribution over $A$ (in $\ell_1$ distance).
	Then, since $p$ is sufficiently close to uniform and since the $N_j$'s are $\eps$-covers for $\eps=O(d/m)$ with respect to the uniform distribution,
	it will follow that $\E_{x\sim p}[c(x)\cdot b(x)]\leq O(d/m)$ for all $b\in \B$, which will show that $\gamma = O(d/m)$ as required.

To construct $c$ and $p$ we consider the polytope defined by the following Linear Program (LP) 
	on variables~$z_{1},\ldots,z_{m}$ with the following constraints: 
\begin{align*}
&-1 \leq z_j \leq +1 \qquad \qquad (\forall j\in [d])\\
&\sum_{i=1}^{m}z_{i}r(x_{i})=0 \qquad \qquad  (\forall j \in [d])~~(\forall r \in N_j)
\end{align*}
Consider a vertex $z=(z_{1},\ldots,z_{m})$ of this polytope.
	 Since the number of equality constraints is at most $m/2$, 
	 there are must be at least $m/2$ inequality constraints that $z$ meets with equality.
	 Namely, $\lvert z_i\rvert =1$ for at least~$m/2$ indices.
	 This implies that $Z:=\|z\|_{1}\ge m/2$. 
	 We assign
labels and probabilities as follows: 
\[
y_{j}=\sign(z_{j}),~~p_{j}=\frac{\lvert z_{j}\rvert}{Z},\quad j=1,\ldots,m.
\]
Let $b\in\mathcal{B}_{j}$. Notice that
\[
\E_{x_i\sim p}[y_i\cdot b(x_i)]=\sum_{i}p_{i}b(x_{i})y_{i}=\frac{1}{Z}\sum_{i}|z_{i}|\,\mathrm{sgn}(z_{i})b(x_{i})=\frac{1}{Z}\sum_{i} z_{i} b(x_{i}).
\]
Pick $r\in N_{j}$ such that $\frac{1}{m}\lvert\{i:~b(x_{i})\neq r(x_{i})\}\rvert\le \frac{40d}{m}$.
Denoting by $I=\{i:~b(x_{i})\neq r(x_{i})\}$ (i.e., $\lvert I\rvert \leq 40d$), the rightmost sum
can be expressed as 
\[
\begin{aligned} & \frac{1}{Z}\sum_{i} z_{i} b(x_{i})=\frac{1}{Z}\sum_{i}z_{i} r(x_{i})+ \frac{1}{Z}\sum_{i\in I} z_{i}(b(x_{i})-r(x_{i}))\\
 & =0+\frac{1}{Z}\sum_{i\in I}z_{i} (b(x_{i})-r(x_{i}))\le \frac{2}{m} \sum_{i\in I} |b(x_{i})-r(x_{i})| =  \frac{4|I|}{m}  \le \frac{160d}{m}
\end{aligned}
\]
Thus, every $b\in\B=\cup_{i\leq d}\B_i$ satisfies $\E_{x_i\sim p}[y_i\cdot b(x_i)] \leq \frac{160d}{m}$,
which implies that $\gamma \leq \frac{160d}{m} = O(d/m)$ (equivalently, $\vc_\gamma(\B)=m = O(d/\gamma)$) as required.
\end{proof}

\subsection{Halfspaces}\label{sec:hs}

For halfspaces in $\mathbb{R}^d$, we give a tight bound on its $\gamma$-VC dimension (in terms of $\gamma$) 
    of $\Theta_d\left(\frac{1}{\gamma}\right)^{\frac{2d}{d+1}}$.
    The upper bound follows from \Cref{thm:ubdisc} and the lower bound is established in the next theorem,
    which also provides a natural condition on a given set of points which implies it can be $\gamma$-interpolated by halfspaces:
\begin{theorem}[\Cref{thm:gammaVCHS} restatement]
Let $\mathsf{HS}_d$ denote the class of halfspaces in $\mathbb{R}^d$ and $\gamma\in(0,1]$. 
Then, 
\[\mathrm{VC}_{\gamma}(\mathsf{HS}_d)=\Theta_d\left(\left (\frac{1}{\gamma}\right)^{\frac{2d}{d+1}} \right).\]
Further, every set $Y\subseteq \R^d$ of size $\Theta_d((\frac{1}{\gamma})^{\frac{2d}{d+1}} )$ is $\gamma$-interpolated by $\mathsf{HS}_d$, provided that $Y$ is \emph{``dense''} in the following sense:
the ratio between the maximal and minimal distances among all distinct pairs of points in $Y$ is bounded by some
$O_d(\lvert Y\rvert^{\frac{1}{d}})$.
\end{theorem}

The proof of \Cref{thm:gammaVCHS} is based on ideas from Discrepancy theory.
	In particular, it relies on the analysis of the discrepancy of halfspaces due to \cite{Alexander90disc}
	(see \cite{matousekDiscrepencyBook} for a text book presentation of this analysis).

\subsubsection{Tools and Notation from Discrepancy Theory}

\paragraph{Weighted Discrepancy.}
Let $p$ a (discrete) distribution over $\X$ and let $c:\X\to\{\pm 1\}$ be a labeling of $\X$ which we think of as a coloring.
	For an hypothesis $b:\X\to\{\pm 1\}$, define the $p$-weighted discrepancy of $c$ with respect to $b$ by
	\[\disc_p(c; b) = \sum_{x : b(x)=1}p(x)\cdot c(x).\]

The following simple identity relates the weighted discrepancy with $\gamma$-realizability.
	For every distribution $p$, target concept $c:\X\to\{\pm 1\}$ and hypothesis $b:\X\to\{\pm 1\}$:
	\begin{equation}\label{eq:4}
	\E_{x\sim p}[c(x)\cdot b(x)] = \disc_p(c;b) - \disc_p(c;-b). 
	\end{equation}

\paragraph{Motion Invariant Measures.}

The proof of \Cref{thm:gammaVCHS} uses a probabilistic argument. 
	In a nutshell, the lower bound on the $\gamma$-VC dimension follows by showing that if $A$ is dense
	then each of its $2^{\lvert A\rvert}$ labelings are $\gamma$-realizable.
	Establishing $\gamma$-realizability is achieved by defining a special distribution $\nu$ over halfspaces such that 
	for every distribution $p$ on $A$ and every labeling $c:A\to\{\pm 1\}$,
	a random halfspace $b\sim \nu$ is $\gamma$-correlated with $c$ with respect to $p$. That is, 
	\[ \E_{b\sim \nu}\Bigl[\E_{x\sim p}[c(x)b(x)]\Bigr]\geq \gamma.\]
	The special distribution $\nu$ over halfspaces which has this property
	is derived from a {\it motion invariant measure}:
	this is a measure over the set of all hyperplanes in $\R^d$ which is invariant under applying rigid motions 
	(i.e., if~$L'$ is a set of hyperplanes obtained by applying a rigid motion on a set $L$ of hyperplanes,
	then the measure of $L$ and $L'$ is the same).
	It can be shown that up to scaling, there is a unique such measure
	(similar to the fact that the Lebesgue measure is the only motion-invariant measure on points in $\R^d$).
	We refer the reader to \cite[Chapter~6.4]{matousekDiscrepencyBook} for more details on how to construct
	this measure and some intuition on how it is used in this context.

One property of this measure that we will use, whose planar version is known by the name {\it the Perimeter Formula},
	is that for any convex set $K$ the set of hyperplanes which intersect $K$ has measure equal to the boundary area of $K$.
	Note that this implies that whenever the boundary area of $K$ is $1$, then this measure
	defines a probability distribution over the set of all hyperplanes intersecting $K$.

\subsubsection{Proof of Theorem~\ref{thm:gammaVCHS}}

The following lemma is the crux of the proof.

\begin{lemma}\label{lem:cruxhs}
Let $\hs_d$ be the class of $d$-dimensional halfspaces. 
Then, every dense set $A$ of size $n$ satisfies
that for every $c:A\to\{\pm 1\}$ and for every distribution $p$ on $A$
there is a halfspace $b\in \hs_d$ such that
\[\disc_p(c;b)=\Omega(n^{-1/2-1/2d}).\]
\label{lem:discHalf} 
\end{lemma}

\Cref{thm:gammaVCHS} is implied by \Cref{lem:cruxhs} as follows:
	let $A$ be a dense set as in $\Cref{lem:cruxhs}$. 
	We need to show that each of the $2^n$ labelings of $A$ are $\gamma$-realizable by $\hs_d$ for $\gamma=\Omega(n^{-1/2-1/2d})$.
	Let $c:A\to\{\pm 1\}$ and let $p$ be a distribution over $A$.
	By \Cref{lem:cruxhs}, there exists $b\in \hs_d$ such that
	\[\disc_p(c;b)\geq \Omega(n^{-1/2-1/2d}).\]
	We distinguish between two cases:
	(i) if $\disc_p(c;-b) \leq 0$, then by \Cref{eq:4}:
	\[\E_{x\sim p}[c(x)\cdot b(x)] = \disc_p(c;b) - \disc_p(c;-b) \geq \Omega(n^{-1/2-1/2d}),\]
	as required. (ii) Else, $\disc_p(c;-b) > 0$ in which case let $b_+$ be a halfspace which contains $A$
	(i.e., $b_+(x)=+1$ for all $x\in A$), and notice that
	\[\E_{x\sim p}[c(x)\cdot b_+(x)] = \disc_p(c;b) + \disc_p(c;-b) \geq \Omega(n^{-1/2-1/2d}).\]
	Thus, in either way there exists a halfspace $b\in \B$ as required.

\begin{proof}
The proof follows along the lines of \cite[Theorem~6.4]{matousekDiscrepencyBook}. 
The main difference is that we consider weighted discrepancy whereas
the proof in \cite{matousekDiscrepencyBook} handles the unweighted
case. We therefore describe the modifications needed to incorporate
weights. 

Following \cite{matousekDiscrepencyBook} we restrict our
attention to the $2$-dimensional case and to sets $A$ which are $n^{1/2}\times n^{1/2}$-regular grids.
The extension of our result
to the general $d$-dimensional case is identical to the extension described
in \cite[page~191]{matousekDiscrepencyBook}.

Let $A\subseteq\mathbb{R}^{2}$ be an $n^{1/2}\times n^{1/2}$ regular grid placed 
	within the square $\mathcal{S}=[0,\frac{1}{4}]^2$.
	Let $c:A\rightarrow\{\pm 1\}$ and $p$ be a distribution over $A$. 
	Our goal is to derive a halfplane $b$ such that $\disc_p(c;b)=\Omega(n^{-1/2-1/2d}) = \Omega(n^{-3/4})$ (as $d=2$).
	The derivation of $b$ is done via a probabilistic argument: that is, we define a distribution $\nu$ over halfplanes
	and show that on average, a halfplane drawn from $\nu$ satisfies the desired inequality.

Following \cite{matousekDiscrepencyBook} denote by $\nu$ a motion-invariant measure on the set of lines which intersect $\mathcal{S}$. 
	Note that $\nu$ is indeed a probability distribution, because the perimeter of $\mathcal{S}$ is $1$.
	By identifying every line with the upper\footnote{We may ignore vertical lines as their $\nu$-measure is $0$.} halfplane it supports,
	we view $\nu$ as a distribution over halfplanes.
	We will prove that
\begin{equation}\label{eq:5}
\sqrt{\E_{b\sim \nu}[D(b)^2]} \geq \Omega(n^{-3/4}),
\end{equation}
where $D(b)=\disc_p(c;b)$. Note that this indeed implies the existence of a halfplane $b$ such that $D(b)\geq \Omega(n^{-3/4})$, as required.

We define the functions $f_{x}:\hs_2\rightarrow\mathbb{R}$, $x\in\mathcal{A}$ as follows. 
Let $I_x:\hs_2 \rightarrow \{0,1\}$ denote the indicator function defined by 
\[I_x(b) = 
\begin{cases}
1	&u(x)=+1\\
0	&u(x)=-1.
\end{cases}
\]
For some sufficiently small constant $\alpha > 0$ (to be determined later),
	let $w=\alpha n^{-1/2}$, and let $\mathsf{w}$ denote the vertical vector $(0,w)$ and let 
\[
f_{x}(b)=I_{x-2\w}(b)-4I_{x-\w}(b)+6I_{x}(b)-4I_{x+\w}(b)+I_{x+2\w}(b).
\]
Define $F(b)=\sum_{x\in A}c(x)f_{x}(b)$. By Cauchy-Schwarz inequality,

\[
\sqrt{\E_{b\sim\nu}[D^2]} \geq \frac{\E_{b\sim\nu}[F\cdot D]}{\sqrt{\E_{b\sim\nu}[F^2]}}.
\]

\Cref{eq:5} follows from bounding $\sqrt{\E[F^{2}]}$ and $\E[F\cdot D]$
from above and from below, respectively. 
The bound 
\begin{equation}\label{eq:6}
\E[F^{2}]= O(\sqrt{n}),
\end{equation}
follows from exactly\footnote{Note that $F$ is defined the same like in \cite{matousekDiscrepencyBook}. The weights only affect the definition of $D$.}
 the same argument as in \cite[pages~190--191]{matousekDiscrepencyBook}.
To bound $\E[F\cdot D]$, note that
\begin{align}
\E[F\cdot D] &= \E_{b\sim \nu}\Bigl[ \Bigl(\sum_x c(x)f_x(b)\Bigr) \Bigl(\sum_{x'} p(x')c(x')I_{x'}(b)\Bigr)\Bigr]\nonumber\\
					&=\sum_{x}p(x)c(x)^{2}\E_b[f_{x}(b)I_{x}(b)]+\sum_{x}\sum_{x'\neq x}p(x)c(x)c(x')\E_b[f_{x}(b)I_{x'}(b)]\nonumber \\
 					& =\sum_{x}p(x)\left(\E_b[f_{x}I_{x}]+\sum_{x'\neq x}c(x)c(x')\E_b[f_{x}I_{x'}]\right) \nonumber\\
					&\ge\sum_{x}p(x)\left(\underbrace{\E_b[f_{x}I_{x}]-\Big|\sum_{x'\neq x}\E_b[f_{x}I_{x'}]\Big|}_{***}\right),\label{eq:7}
\end{align}
where in the last inequality we used that $\lvert c(x)\rvert = 1$ for all $x\in A$.
The following calculations are derived in \cite[pages~190--191]{matousekDiscrepencyBook}
(recall that $w=\alpha n^{-1/2}$ where $\alpha$ is a sufficiently small constant): 
\begin{itemize}
\item for any $x\in A$, 
\[
\E[f_{x}I_{x}]=4w=4\alpha n^{-1/2},
\]
\item for any $x\in A$, 
\[\Big|\sum_{x'\neq x}\E[f_{x}I_{x'}]\Big| = O(n^{3/2} w^4) = O(\alpha^4 n^{-1/2})\]
\end{itemize}
Thus, by taking $\alpha$ to be sufficiently small, the term ({*}{*}{*})
in Equation \ref{eq:7} is lower bounded by $\Omega(n^{-1/2})$.
Since $\sum p(x)=1$ it follows that also 
\begin{equation}\label{eq:8}
\E[F\cdot D]=\Omega(n^{-1/2}).
\end{equation}
All in all, \Cref{eq:6,eq:8} imply that
\[
\sqrt{\E [D^{2}]}\geq \frac{\E[F\cdot D]}{\sqrt{\E[F^{2}]}}=\Omega\Bigl(\frac{n^{-1/2}}{n^{1/4}}\Bigr)=\Omega(n^{-3/4}),
\]
which establishes \Cref{eq:5} and finishes the proof.
\end{proof}

\section{Conclusion and Open Problems}\label{sec:conc}

We conclude the paper with some suggestions for future research:
\begin{itemize}
\item \Cref{alg:sepBoost} suggests a possibility of improved boosting algorithms which exploit the simplicity of the base-class
and use more complex (``deeper'') aggregation rules. It will be interesting to explore efficient realizations of \Cref{alg:sepBoost},
for realistic base-classes $\B$.
\item The bounds provided on the $\gamma$-VC dimensions of halfspaces and decision stumps are rather loose in terms of $d$.
It will be interesting to find tight bounds. Also, it will be interesting to explore how the $\gamma$-VC dimension behaves under natural operations.
For example, for $k>0$ consider the class $\B'$ of all $k$-wise majority votes of hypotheses from $\B$. How does $\vc_\gamma(\B')$ behaves as a function of $k$ and $\vc_\gamma(\B)$?
\item Characterize for which classes $\B$ there exist boosting algorithms which output weighted majorities of the base hypotheses using much less than $\tilde O(\gamma^{-2})$ oracle calls; e.g., for which classes is it possible to use only~$\tilde O(\gamma^{-1})$ oracle calls?
\end{itemize}

\section*{Acknowledgements}
We thank Yoav Freund and Rob Schapire for useful discussions.
We also thank the anonymous reviewers of previous versions of this manuscript and helped improving it.
In particular, we thank one of the reviewers for suggesting the last open question in Section~\ref{sec:conc}.

\printbibliography
\end{document}